\documentclass[lettersize,journal]{IEEEtran}
\usepackage{amsmath,amsfonts}
\usepackage{algorithmic}
\usepackage{algorithm}
\usepackage{array}
\usepackage{textcomp}
\usepackage{stfloats}
\usepackage{url}
\usepackage{verbatim}
\usepackage{graphicx}
\usepackage{subcaption}
\usepackage{cite}
\usepackage{booktabs} % for professional tables
\usepackage{multirow}
\usepackage[table]{xcolor}
\usepackage{xspace}
\newcommand{\abb}{\texttt{InA-Probe}\xspace}
\usepackage[most]{tcolorbox}

\usepackage{amsthm}

\theoremstyle{definition}

\newtcolorbox{promptbox}[1][]{
    enhanced,
    breakable,
    colback=gray!5,
    colframe=blue!60!black,
    arc=4pt,
    boxrule=1pt,
    left=6pt, right=6pt, top=6pt, bottom=6pt,
    fonttitle=\bfseries\sffamily\small,
    title=#1,
    before skip=1em,
    after skip=1em,
    boxsep=0pt,
    sharp corners=northwest,
    attach boxed title to top left={yshift=-2pt, xshift=4pt},
    boxed title style={
        colback=blue!60!black,
        colframe=blue!60!black,
        sharp corners,
        rounded corners=northwest,
        top=2pt,      % 上内边距
        bottom=2pt,   % 下内边距
    }
}

\newcommand{\placeholder}[1]{\textcolor{red!70!black}{\texttt{#1}}}

\newtcolorbox{taskinstructionbox}[1][]{
    enhanced,
    breakable,
    colback=black!2,               % 极浅灰背景
    colframe=teal!60!black,        % 深青色边框
    arc=4pt,
    boxrule=1pt,
    left=3pt, right=1pt, top=8pt, bottom=8pt,
    fontupper=\ttfamily\small,
    before skip=1em,
    after skip=1em,
    title=#1, 
    fonttitle=\bfseries\sffamily\small,
    coltitle=white,
    attach boxed title to top left={yshift=-4pt, xshift=4pt},
    boxed title style={colback=teal!70!black, sharp corners}
}

\begin{document}

% \title{Beyond Passive Alignment: Instruction-Aware Active Probing for Time Series Forecasting with LLMs}
\title{InA-Probe: Instruction-Aware Active Probing for Time Series Forecasting with LLMs}

\author{Peiliang Gong, Emadeldeen Eldele, Chenyu Liu, Ziyu Jia, Yi Ding, Xinliang Zhou, Lianchao Gu, Qi Zhu, Yang Liu, ~\IEEEmembership{Senior Member,~IEEE,} Daoqiang Zhang, ~\IEEEmembership{Senior Member,~IEEE,} and Xiaoli Li, ~\IEEEmembership{Fellow,~IEEE}
\thanks{Peiliang Gong, Chenyu Liu, Yi Ding, Xinliang Zhou, and Yang Liu are with the 
College of Computing and Data Science, Nanyang Technological University, Singapore.}
\thanks{Emadeldeen Eldel is with the Department of Computer Science, Khalifa University, UAE.}
\thanks{Lianchao Gu, Qi Zhu and Daoqiang Zhang are with the Key Laboratory of Brain-Machine Intelligence Technology, Ministry of Education, College of Artificial Intelligence, Nanjing University of Aeronautics and Astronautics, Nanjing 211106, China.}
\thanks{Xiaoli Li is with the Information Systems Technology and Design, Singapore University of Technology and Design, Singapore.}
} % end here.

\markboth{Journal of \LaTeX\ Class Files,~Vol.~14, No.~8, August~2021}%
{Shell \MakeLowercase{\textit{et al.}}: A Sample Article Using IEEEtran.cls for IEEE Journals}


\maketitle

\begin{abstract}
Large Language Models (LLMs) have recently demonstrated impressive potential for time series forecasting. However, existing methods predominantly rely on passive modality alignment or static task reprogramming, which often fail to capture fine-grained, non-stationary temporal patterns or to adapt to nuanced task intents. In this paper, we propose \textbf{In}struction-aware \textbf{A}ctive \textbf{Prob}ing (\abb), which shifts the paradigm from passive alignment toward an active, instruction-driven probing mechanism. Specifically, we design a Multi-Level Instruction Injection mechanism that enriches the model with both global task objectives and fine-grained, patch-level semantic priors. Building on this, an Adaptive Query Generation module produces sample-specific probes that are dynamically modulated by the temporal context. These probes are then refined through a dual-stage attention process: they first internalize task-specific intents via Instruction-Aware Self-Attention, and subsequently interrogate the projected temporal representations through Temporal Cross-Attention to extract salient patterns. Comprehensive experiments on seven real-world benchmarks show that \abb consistently outperforms state-of-the-art deep learning and LLM-based baselines, excelling in both one-for-all generalization and zero-shot transfer while reducing forecasting error by up to 37\% in challenging cross-domain scenarios. Ablation studies further confirm that the synergy between adaptive querying and fine-grained instructions is key to unlocking the reasoning power of LLMs for complex time series.
\end{abstract}


\section{Introduction}
\IEEEPARstart{A}{ccurate} long-term time series forecasting (LTSF) serves as a fundamental pillar for strategic decision-making in diverse critical domains, ranging from power grid load management \cite{gasparin2022deep} and urban traffic control \cite{chen2022bidirectional} to precision meteorology. In recent years, the research paradigm has decisively shifted from classical statistical methodologies toward deep-learning-based architectures, which have redefined the boundaries of predictive performance. State-of-the-art specialized forecasters, such as iTransformer \cite{LiuHZWWML24}, PatchTST \cite{NieNSK23}, and TimesNet \cite{WuHLZ0L23}, have demonstrated exceptional capabilities in distilling complex temporal dependencies and local structural patterns from large-scale historical observations, establishing strong benchmarks for reliable operational planning.

\begin{figure}[t]
  % \vskip 0.2in
  \begin{center}
    \centerline{\includegraphics[width=\columnwidth]{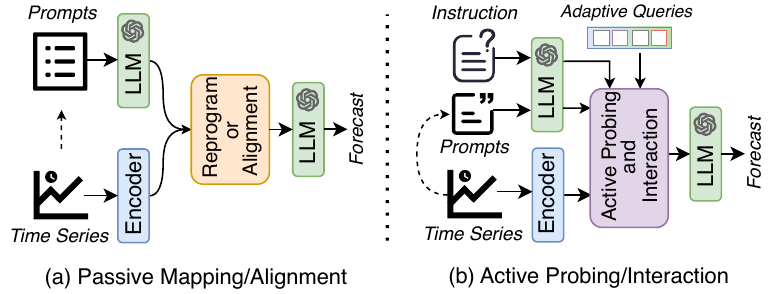}}
    \caption{
      Paradigm shift from passive alignment to active instruction-aware probing. (a) Existing methods rely on reprogramming or passive alignment. (b) Ours enables active, instruction-driven probing by synthesizing sample-specific queries that internalize task intents to dynamically interrogate temporal features.
    }
    \label{fig:motivation}
    % \vspace{-0.7cm}  % 关键！压缩图与正文间隙
  \end{center}
\end{figure}

Parallel to the advancement of specialized architectures, a transformative trend has emerged that treats time series data as a sequence of linguistic tokens, leveraging the profound sequence-modeling capabilities of pre-trained Large Language Models (LLMs). This \textit{``Time-as-Language"} paradigm posits that the universal reasoning priors embedded within frozen LLM backbones can be effectively repurposed for temporal forecasting, even across disparate domains. By preserving the pre-trained weights of models such as GPT-2 \cite{radford2019language} or Llama \cite{touvron2023llama}, emerging frameworks, exemplified by TimeLLM \cite{0005WMCZSCLLPW24} and AutoTimes \cite{liu2024autotimes}, demonstrate remarkable adaptability and competitive performance with minimal task-specific overhead. These LLM-based forecasters capitalize on the high-dimensional feature spaces learned during linguistic pre-training to recognize fundamental temporal structures, offering a promising route toward more unified and general-purpose forecasters.

Despite their promise, existing LLM-based forecasters are often hindered by a \textit{passive processing} bottleneck, as illustrated in Figure~\ref{fig:motivation}(a). Temporal patches are mapped unidirectionally into the linguistic latent space through coarse-grained alignment~\cite{liu2024autotimes, sun2025adapting} or task reprogramming~\cite{0005WMCZSCLLPW24}, with no mechanism for the model to selectively query the encoded representations according to the task or the local signal regime. The frozen backbone thus acts as a passive recipient rather than an active reasoner. It cannot ground numerical fluctuations in the physical concepts they represent, which limits its ability to track non-stationary dynamics and adapt to differing task intents, precisely the settings where its pre-trained reasoning priors should help most.

To bridge this gap, we propose a paradigm shift from passive alignment to active instruction-aware probing, as depicted in Figure~\ref{fig:motivation}(b). We introduce \abb, which recasts the LLM from a passive recipient of pre-processed embeddings into an active probing agent. Rather than projecting temporal patches blindly, \abb synthesizes \textit{sample-specific queries} that first internalize the task intent from linguistic instructions and then interrogate the encoded temporal representations for semantically salient patterns. This instruction-conditioned interrogation grounds the LLM's pre-trained reasoning in the physical dynamics of the signal, letting it attend to the patterns that matter for the current task and window.

The technical efficacy of \abb stems from three synergistic components designed to navigate the interplay between heterogeneous modalities. First, the Adaptive Query Generation (AQG) module distills local temporal context into sample-specific probes, which are dynamically modulated by global task intents and instance-wise importance gating. These probes subsequently enter the Time-Language (TL-) Connector, where an instruction-aware self-attention mechanism internalizes the semantic anchors of the global prompts, followed by a temporal cross-attention layer that selectively interrogates the projected temporal representations to locate salient patterns. To further fortify the link between numerical signals and linguistic reasoning, we also adopt a Patch-wise Contrastive Alignment during the training phase. This objective utilizes dense, segment-level semantic descriptors to provide fine-grained supervision, ensuring that the latent representation space is fundamentally grounded in the physical attributes of the time series. By integrating these modules, the framework transforms the frozen LLM into a context-aware forecaster capable of precise, generative reasoning.
The contributions of this paper can be summarized as follows:
\begin{itemize}
    \item We propose \abb, a novel framework that shifts the paradigm of LLM-based forecasting from passive sequence processing to instruction-aware active probing, effectively bridging the structural and semantic gap between numerical time series and linguistic reasoning.
    \item We design a specialized TL-Connector that internalizes task instructions to drive targeted feature interrogation. In addition, we propose an AQG module to enable sample-aware probing and a patch-wise contrastive alignment objective to provide dense, semantic-level supervision for numerical-linguistic grounding.
    \item \abb demonstrates superior performance against state-of-the-art deep learning and LLM-based baselines across seven diverse benchmarks, achieving significant gains in both one-for-all generalization and zero-shot transfer scenarios while maintaining an optimized computational and memory footprint.
\end{itemize}

\section{Related Work}
\subsection{Deep Learning for Time Series Forecasting}
The landscape of LTSF has undergone a significant transformation, evolving from classical statistical methods to sophisticated deep learning architectures capable of capturing intricate temporal dependencies \cite{kong2025deep, wang2024deep, lim2021time, han2019review}. Early deep learning efforts focused on Recurrent Neural Networks (RNNs) \cite{lai2018modeling, salinas2020deepar} and Convolutional Neural Networks (CNNs) \cite{WuHLZ0L23, Eldele0C0024}; however, the introduction of the Transformer architecture marked a pivotal shift due to its superior capacity for modeling long-range interactions \cite{li2019enhancing}. To address the quadratic computational complexity of standard self-attention in LTSF, numerous variants emerged, such as Informer \cite{zhou2021informer}, Autoformer \cite{wu2021autoformer}, and FEDformer \cite{zhou2022fedformer}, which utilized sparse attention mechanisms and decomposition techniques to process extended sequences.
Despite these advancements, the necessity of complex attention mechanisms was challenged by the emergence of DLinear \cite{zeng2023transformers}, which demonstrated that a simple decomposition-based linear model could outperform many Transformer-based forecasters. This sparked a renewed focus on structural efficiency and local semantic modeling. PatchTST \cite{NieNSK23} subsequently redefined the state-of-the-art by introducing a patching mechanism that aggregates local temporal information into sub-series level tokens, effectively enhancing both forecasting accuracy and computational throughput. More recent innovations, such as iTransformer \cite{LiuHZWWML24}, further improved performance by inverting the traditional time-series dimensions to capture global multivariate correlations within the embedding space. While these models have achieved remarkable numerical precision, they are primarily designed as task-specific forecasters that rely on passive numerical mapping. 
% Consequently, they often lack the cross-domain reasoning and semantic grounding required to adapt to diverse task instructions or zero-shot scenarios, establishing a clear need for more generalized and instruction-aware paradigms.

\subsection{Large Language Models for Time Series}
The unprecedented success of LLMs in natural language processing has catalyzed a burgeoning research direction that repurposes their universal reasoning priors for temporal data analysis \cite{cao2024tempo, pan2024s, hu2025contextalignment}. This \textit{``Time-as-Language"} paradigm posits that time series, when properly tokenized, can be treated as a specialized sequence modality amenable to the self-attention mechanisms of frozen backbones \cite{gruver2023large}. Early explorations, such as FPT \cite{zhou2023one} and Timer \cite{LiuZLH0L24}, demonstrated that LLMs possess inherent zero-shot capabilities, allowing them to recognize fundamental temporal patterns across diverse domains without extensive task-specific fine-tuning. Current state-of-the-art methodologies primarily revolve around passive modality alignment, where temporal patches are reprogrammed or projected into the linguistic latent space \cite{0005ZCZL0WPW24, zhang2024large, liu2025calf, liu2025timecma}.
Frameworks like TimeLLM \cite{0005WMCZSCLLPW24} and AutoTimes \cite{liu2024autotimes} utilize prefix prompting and linear alignment to instruct the LLM to process numerical sequences. While effective, these approaches typically adopt a unidirectional mapping strategy, treating the LLM as a passive recipient of pre-processed embeddings. Consequently, the model lacks a mechanism to dynamically interrogate the encoded temporal representations based on the specific nuances of a task or the non-stationary properties of a signal. Although recent works such as TALON~\cite{sun2025adapting} have begun incorporating natural language instructions to guide forecasting, they still largely rely on passive alignment, leaving the LLM's reasoning capacity underused for actively probing salient physical patterns.
Distinct from this frozen-LLM line, a parallel direction trains \emph{time-series-native} foundation models from scratch on large corpora, such as TimesFM~\cite{das2024timesfm}, Moirai~\cite{woo2024moirai}, and Chronos~\cite{AnsariSTZMSSRPK24}. Our work instead repurposes a frozen language model and focuses on \emph{how} its pre-trained reasoning is accessed, through active, instruction-aware probing rather than pre-training a new temporal backbone.

\section{Method}
\subsection{Problem Definition}
Given a historical lookback window $\mathbf{X} = \{\mathbf{x}_1, \dots, \mathbf{x}_L\} \in \mathbb{R}^{L \times C}$, where $L$ denotes the lookback length and $C$ represents the number of variables, the objective of time series forecasting is to predict the subsequent values $\mathbf{Y} = \{\mathbf{x}_{L+1}, \dots, \mathbf{x}_{L+T}\} \in \mathbb{R}^{T \times C}$ over a future horizon $T$. Formally, we aim to learn a predictive mapping $\mathcal{F}: \mathbf{X} \rightarrow \mathbf{Y}$ that minimizes the forecast error. In our proposed framework, this mapping is augmented by incorporating multi-level linguistic instructions $\mathcal{I}$ to guide the model in capturing non-stationary temporal dynamics, effectively learning the mapping $\hat{\mathbf{Y}} = \mathcal{F}(\mathbf{X}, \mathcal{I})$.

\begin{figure*}[ht]
  % \vskip 0.2in
  \begin{center}
    \centerline{\includegraphics[width=1.0 \textwidth]{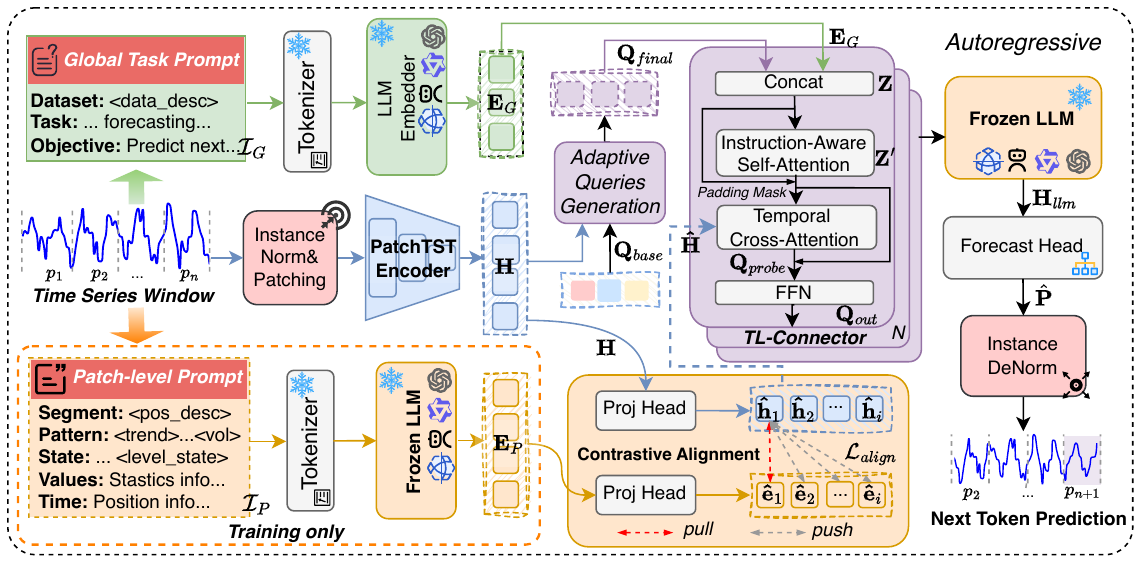}}
    \caption{
      Framework of the \abb. The pipeline illustrates the transition from raw numerical signals to generative forecasting. Input time series are processed into patch embeddings and interrogated by Adaptive Queries within the Time-Language Connector. This connector leverages instruction-aware self-attention to internalize Global Task Prompts and temporal cross-attention to extract salient features. During the training phase, Patch-level Prompts provide dense semantic supervision via a contrastive alignment objective to bridge the modality gap, while the Frozen LLM remains unchanged to preserve its pre-trained reasoning priors.
    }
    \label{fig:framework}
    \vspace{-0.4cm}
  \end{center}
\end{figure*}

\subsection{Overview}
We propose \abb, an instruction-aware active probing framework designed to shift LTSF from passive alignment to a task-oriented probing paradigm, as shown in Figure \ref{fig:framework}. The architecture integrates three core modules: a Multi-Level Instruction Injection mechanism that enriches the model with both global task objectives and patch-level semantic priors; an AQG module that turns a set of learnable \emph{query} vectors, which serve as the probing operators of our framework, into sample-specific probes dynamically modulated by the temporal context; and an Instruction-Aware Probing process that internalizes task intents through self-attention and actively interrogates the encoded temporal representations via cross-attention. To further bridge the modality gap, we incorporate Patch-wise Contrastive Alignment to explicitly supervise the correspondence between temporal patterns and their linguistic descriptions. The details of our method are described below.

\subsection{Time Series Encoding} Following the channel-independent setting, each univariate series is modeled separately. Given a lookback window $\mathbf{X}$, we first apply instance normalization~\cite{KimKTPCC22} to mitigate distribution shift, then segment the normalized series into $N$ non-overlapping patches of length $P$. A PatchTST-style encoder~\cite{NieNSK23} maps these patches into latent representations $\mathbf{H}=\{\mathbf{h}_1,\dots,\mathbf{h}_N\}\in\mathbb{R}^{N\times D}$, where $N$ is the number of patches and $D$ the embedding dimension.

\subsection{Multi-Level Instruction Injection}
To bridge the gap between numerical signals and semantic reasoning, we inject linguistic instructions at two complementary granularities: a \emph{global} prompt that conveys the task- and domain-level context, and \emph{patch-level} prompts that describe the local dynamics of each segment. This hierarchical design lets the model perceive the forecasting task both as a whole and at the resolution of individual patches.

\subsubsection{\textbf{Global Task Instruction}}
The global instruction $\mathcal{I}_G$ conveys the overarching context of the forecasting task, comprising a dataset-specific description (\textit{e.g., domain characteristics}), the forecasting horizon, and the high-level objective. Embedding these global priors makes $\mathcal{I}_G$ a semantic anchor for the model's understanding of the data distribution. The template is shown below. The full dataset-specific descriptions are provided in the Appendix.

\begin{taskinstructionbox}[Global Task Instruction]
\textbf{Dataset}: \placeholder{<dataset\_desc>}. \\
\textbf{Task}: Long term time series forecasting. \\
\textbf{Objective}: Predict the next \placeholder{<token\_len>} timesteps based on historical patterns.
\end{taskinstructionbox}

\subsubsection{\textbf{Patch-level Hybrid Prompt}}
To capture localized non-stationary dynamics, we attach one hybrid prompt to each
temporal patch, yielding $\mathcal{I}_P = \{i_{p,1}, \dots, i_{p,N}\}$. For patch $p_n$, the prompt $i_{p,n}$ encodes four complementary semantic dimensions: (a) \textit{Temporal Position}, an indicator (\textit{e.g.}, Start, Mid, End) of the patch's location within the lookback window; (b) \textit{Pattern Semantics}, qualitative descriptors of the signal's shape, covering trend direction (\textit{e.g.}, Steady Rise, Sharp Fall) and volatility (\textit{e.g.}, High Volatility, Smooth); (c) \textit{State Level}, a
categorical indicator of the relative magnitude of the values (\textit{e.g.}, Extreme High, Low Level); and (d) \textit{Value Statistics}, quantitative descriptors such as the local slope and value range that ground the qualitative labels in numerical precision. These fields are assembled into the structured template below.

\begin{taskinstructionbox}[Patch-level Hybrid Prompt]
\textbf{Segment}: \placeholder{<pos\_desc>}. \\
\textbf{Pattern}: \placeholder{<trend>} with \placeholder{<vol>} \placeholder{<peak>}. \\
\textbf{State}: Currently at \placeholder{<level>}. \\
\textbf{Values}: range \placeholder{<r\_min>} to \placeholder{<r\_max>}, slope \placeholder{<slope\_value>}. \\
\textbf{Time}: From \placeholder{<x\_p\_start>} to \placeholder{<x\_p\_end>}.
\end{taskinstructionbox}

All descriptors are computed from the $Z$-score normalized patch values, so that local
fluctuations are interpreted relative to the global distribution of the dataset and the
decision thresholds below remain comparable across datasets.
\begin{itemize}
    \item \textit{Trend and shape.} The local trend is obtained by fitting a first-order
    polynomial to the patch values on normalized coordinates $x \in [0,1]$, which makes
    the slope $\theta$ independent of patch length: \emph{Sharp Rise/Fall}
    ($|\theta| > 0.5$), \emph{Steady Rise/Fall} ($0.1 < |\theta| \leq 0.5$), and
    \emph{Flat} ($|\theta| \leq 0.1$).
    \item \textit{Volatility and peaks.} The volatility $\sigma_{local}$ is the standard
    deviation of the patch; since the data is $Z$-score normalized, $\sigma_{local} > 1$
    means the local fluctuation exceeds the global variance: \emph{High Volatility}
    ($\sigma_{local} > 1.0$), \emph{Active Fluctuation} ($0.5 < \sigma_{local} \leq 1.0$),
    and \emph{Smooth} ($\sigma_{local} \leq 0.5$). A patch whose extremum exceeds
    $\pm 2.5$ standard deviations is additionally flagged as \emph{Hits Global
    Peak/Bottom}.
    \item \textit{State and value grounding.} The \emph{State} field reports a relative
    ``water level'' from the patch mean $\mu_{patch}$, ranging from \emph{Extreme Low}
    ($\mu_{patch} < -1.5$) to \emph{Extreme High} ($\mu_{patch} > 1.5$). The \emph{Values}
    and \emph{Time} fields record the raw numerical range and physical time span, keeping
    the descriptors anchored to the actual signal boundaries.
\end{itemize}
As detailed in Section~\ref{sec:opt_inference}, these patch-level prompts supervise the
encoder during training and are not required at inference.

\begin{figure}[t]
  % \vskip 0.2in
  \begin{center}
    \centerline{\includegraphics[width=1.0 \columnwidth]{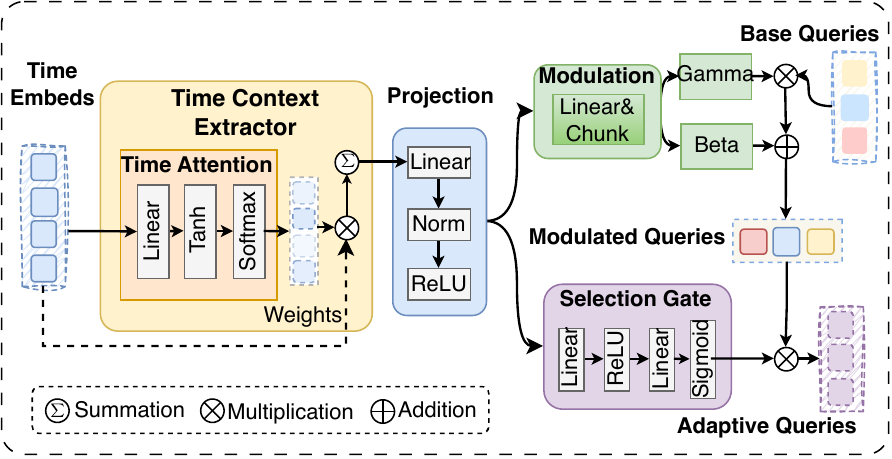}}
    \caption{
      Framework of the AQG module. A time-context extractor with attention pooling distills a global representation from the input embeddings, which then drives two paths, FiLM-based modulation and instance-wise selection gating, that turn the learnable base queries into sample-specific adaptive probes.
    }
    \label{fig:adaptiveQuery}
    % \vspace{-0.4cm}
  \end{center}
\end{figure}

\subsubsection{\textbf{Instruction Embedding}}
Both global and patch-level instructions are processed through a tokenizer and a frozen
LLM embedder $\Phi_{LLM}$. For the global prompt we retain the full token sequence,
whereas for each patch-level prompt we take the embedding of its final token as a single
descriptor:
$$\mathbf{E}_G = \Phi_{LLM}(\mathcal{I}_G)\in\mathbb{R}^{L_g\times D}, \qquad
\mathbf{E}_P = \Phi_{LLM}(\mathcal{I}_P)\in\mathbb{R}^{N\times D},$$
where $L_g$ is the global-prompt length and $D$ the embedding dimension. Thus
$\mathbf{E}_P$ holds one descriptor per patch, matching the $N$ patch embeddings in
$\mathbf{H}$, while $\mathbf{E}_G$ provides the primary guidance for the subsequent
adaptive probing process.

\subsection{Adaptive Query Generation}
% NEW
At the core of our active probing mechanism is a set of \emph{queries}: learnable vectors that act as probing operators, later used to interrogate the projected temporal representations via attention (Section~\ref{sec:probing}). A naive instantiation uses \emph{static} queries, i.e., a single set of vectors that is shared across all input windows and held fixed after training. Such queries impose an identical interrogation strategy on every sample and therefore cannot adapt to the non-stationary dynamics that differ from one window to the next. To overcome this limitation, the AQG module conditions the queries on each input, modulating learnable query prototypes by the current temporal context to produce sample-specific probes. As illustrated in Figure \ref{fig:adaptiveQuery}, the AQG process mainly involves three stages.

\subsubsection{\textbf{Time Context Extraction}}
We first summarize the window into a context vector by attention pooling over the encoded patches: 
$$s_n = \mathrm{Tanh}(\mathrm{Linear}(\mathbf{h}_n)),$$
$$\boldsymbol{\alpha} = \mathrm{Softmax}(s_1,\dots,s_N), \quad \mathbf{c} = \sum_{n=1}^{N}\alpha_n \mathbf{h}_n,$$ and refine it through a projection to obtain the modulation context: $$\mathbf{z} = \mathrm{ReLU}\big(\mathrm{LN}(\mathrm{Linear}(\mathbf{c}))\big) \in \mathbb{R}^{D},$$ which summarizes the global statistical properties and evolutionary dynamics of the window.

\subsubsection{\textbf{Dynamic Modulation}}
We further introduce a set of learnable base query prototypes $\mathbf{Q}_{base} \in \mathbb{R}^{K \times D}$. We instantiate one base query per patch, i.e., $K=N$, so that each probe is tied to a temporal position and the resulting tokens preserve the ordering of the series. To render these queries sample-aware, we utilize a feature-wise linear modulation mechanism that generates scaling and shifting factors derived from the context vector $\mathbf{z}$:
$$[\gamma, \beta] = \text{Split}(\text{Linear}(\mathbf{z})), \mathbf{Q}_{mod} = (1 + \gamma) \odot \mathbf{Q}_{base} + \beta,$$
where $\gamma,\beta\in\mathbb{R}^{D}$ are broadcast across the $K$ prototypes.
Through this modulation, the base queries are projected into a specialized feature space that aligns with the current data distribution.

\subsubsection{\textbf{Selection Gating}}
Recognizing that not all query prototypes are equally relevant to every sample, we further incorporate an instance-wise selection gating mechanism. This gate adaptively weights the modulated queries based on the refined temporal context:
$$\mathbf{g} = \sigma(\text{MLP}(\mathbf{z})) \in \mathbb{R}^{K}, \mathbf{Q}_{final} = \mathbf{Q}_{mod} \odot \mathbf{g}$$
where $\sigma$ denotes the sigmoid function and $\mathbf{Q}_{final} \in \mathbb{R}^{K \times D}$ represents the final set of adaptive queries. This multi-level modulation ensures that the probes can flexibly adjust their focus, such as prioritizing sudden spikes or stable trends, providing a robust foundation for the subsequent instruction-driven probing.

\subsection{Instruction-Aware Active Probing}
\label{sec:probing}
To realize the shift from passive alignment to active feature discovery, we introduce the Instruction-Aware Active Probing mechanism, implemented as the TL-Connector. Unlike standard cross-modal projectors that map temporal features into the language space in a single feed-forward step, the connector follows a query-based cross-modal attention design~\cite{li2023blip,dai2023instructblip}, which we adapt to time series with sample-specific probes from the AQG module and a patch-wise contrastive objective. The connector stacks $N_{conn}$ \emph{Instruct-Probing Blocks}, each applying the dual-stage attention strategy below. For clarity we describe a single block, taking $\mathbf{Q}_{final}$ as its query input.

\subsubsection{\textbf{Instruction-Aware Self-Attention}}
In the first stage, the adaptive queries $\mathbf{Q}_{final}$ are concatenated with the global instruction embeddings $\mathbf{E}_G$ into a joint sequence, over which we apply self-attention so that the queries perceive and internalize the overarching task intent: $$\mathbf{Z} = [\mathbf{Q}_{final} \, ; \, \mathbf{E}_G] \in \mathbb{R}^{(K+L_g) \times D},$$ $$\mathbf{Z}' = \text{LayerNorm}(\mathbf{Z} + \text{Self-Attn}(\mathbf{Z}, \mathbf{M}_{mask})),$$ where $\mathbf{M}_{mask}$ is a padding mask over the instruction tokens that lets the queries attend only to valid semantic information. The self-attention updates the joint sequence as a whole. We then take the first $K$ rows of $\mathbf{Z}'$, the positions corresponding to the queries, as the instruction-aware queries: $$\mathbf{Q}_{inst} = \mathbf{Z}'_{1:K} \in \mathbb{R}^{K \times D}.$$ That is, $\mathbf{Q}_{inst}$ is the query sub-block of $\mathbf{Z}'$ rather than the full sequence. The remaining $L_g$ rows (the instruction tokens) have served their purpose as semantic context and are discarded. Through this interaction, $\mathbf{Q}_{inst}$ absorbs the global task intent carried by $\mathbf{E}_G$.

\subsubsection{\textbf{Temporal Cross-Attention}}
We first map $\mathbf{H}$ through a projection $g_t(\cdot)$, $\hat{\mathbf{H}} = g_t(\mathbf{H}) \in \mathbb{R}^{N \times D}$, which provides the keys and values. Here, $g_t$ reduces to the identity when the encoder and connector share the same width. The cross-attention then lets each query selectively aggregate the temporal patterns relevant to its internalized intent: $$\mathbf{Q}_{probe} = \text{LayerNorm}(\mathbf{Q}_{inst} + \text{Cross-Attn}(\mathbf{Q}_{inst}, \hat{\mathbf{H}}, \hat{\mathbf{H}})).$$ The same features $\hat{\mathbf{H}}$ are aligned with the linguistic descriptors by the patch-wise contrastive objective, so the queries probe a representation that has been grounded against those descriptors rather than pooling all patches uniformly.

\subsubsection{\textbf{Feature Refinement and Output}}
Finally, the probed features are refined by a Feed-Forward Network (FFN): $$\mathbf{Q}_{out} = \text{LayerNorm}(\mathbf{Q}_{probe} + \text{FFN}(\mathbf{Q}_{probe})).$$ The block output $\mathbf{Q}_{out}$ is passed to the next Instruct-Probing Block as its query input. After the final block, $\mathbf{Q}_{out}$ is projected to match the input dimension of the frozen LLM. By decoupling semantic internalization from temporal interrogation, the connector produces representations that are not merely projected into the language space but explicitly conditioned on the task instructions.

\subsection{Generative Decoding and Forecasting}
\label{sec:generative_decoding_forecasting}
The final stage turns the instruction-aware representations into numerical forecasts, using the frozen LLM as a sequence model.

\begin{table*}[ht]
\caption{Multivariate long-term forecasting results under the one-for-all protocol. A single model is trained on a 96-step horizon and recursively applied to all other prediction lengths. The ``$1^{st}$ Count" indicates the total number of best-performing metrics (MSE and MAE) achieved by each model. \textcolor{blue}{\textbf{Blue}} and \textcolor{purple}{\underline{Purple}} denote the best and second-best results, respectively. Full results are provided in Appendix.}
\label{tab:Rolling Forecast Results}
\centering
\renewcommand{\arraystretch}{1.3} % 稍微增加行高，避免文字紧贴横线
\setlength{\tabcolsep}{5.5pt}      % 紧凑列间距，确保25列能塞进页面
\resizebox{\textwidth}{!}{
\begin{tabular}{l | cc | cc | cc | cc | cc | cc | cc | cc | cc | cc | cc | cc}
\toprule
\multirow{3}{*}{\textbf{Model}} & \multicolumn{12}{c|}{\textbf{LLM-based methods}} & \multicolumn{12}{c}{\textbf{Deep learning forecasting methods}} \\
\cmidrule(lr){2-13} \cmidrule(lr){14-25}
& \multicolumn{2}{c|}{{\abb}} & \multicolumn{2}{c|}{TALON} & \multicolumn{2}{c|}{CALF} & \multicolumn{2}{c|}{AutoTimes} & \multicolumn{2}{c|}{TimeLLM} & \multicolumn{2}{c|}{FPT} & \multicolumn{2}{c|}{SimpleTM} & \multicolumn{2}{c|}{Timer\_XL} & \multicolumn{2}{c|}{TimeMixer} & \multicolumn{2}{c|}{iTransformer} & \multicolumn{2}{c|}{PatchTST} & \multicolumn{2}{c}{TimesNet} \\
& \multicolumn{2}{c|}{{(Ours)}} & \multicolumn{2}{c|}{\cite{sun2025adapting}} & \multicolumn{2}{c|}{\cite{liu2025calf}} & \multicolumn{2}{c|}{\cite{liu2024autotimes}} & \multicolumn{2}{c|}{\cite{0005WMCZSCLLPW24}} & \multicolumn{2}{c|}{\cite{zhou2023one}} & \multicolumn{2}{c|}{\cite{ChenLMS25}} & \multicolumn{2}{c|}{\cite{LiuQ00L25}} & \multicolumn{2}{c|}{\cite{WangWSHLMZ024}} & \multicolumn{2}{c|}{\cite{LiuHZWWML24}} & \multicolumn{2}{c|}{\cite{NieNSK23}} & \multicolumn{2}{c}{\cite{WuHLZ0L23}} \\
\midrule
Metric  & MSE & MAE & MSE & MAE & MSE & MAE & MSE & MAE & MSE & MAE & MSE & MAE & MSE & MAE & MSE & MAE & MSE & MAE & MSE & MAE & MSE & MAE & MSE & MAE \\
\midrule
ETTh1   & \textcolor{purple}{\underline{0.395}} & \textcolor{blue}{\textbf{0.417}} & \textcolor{blue}{\textbf{0.386}} & \textcolor{purple}{\underline{0.420}} & 0.416 & 0.429 & {0.402} & 0.428 & 0.542 & 0.520 & 0.422 & 0.437 & 0.424 & 0.450 & 0.407 & 0.429 & 0.418 & 0.434 & 0.432 & 0.451 & 0.441 & 0.451 & 0.495 & 0.489 \\
ETTh2   & \textcolor{blue}{\textbf{0.338}} & \textcolor{blue}{\textbf{0.386}} & \textcolor{purple}{\underline{0.355}} & \textcolor{purple}{\underline{0.395}} & 0.373 & 0.419 & 0.400 & 0.431 & 0.416 & 0.446 & 0.370 & 0.407 & 0.367 & 0.414 & 0.377 & 0.414 & 0.385 & 0.417 & 0.399 & 0.423 & 0.392 & 0.429 & 0.455 & 0.463 \\
ETTm1   & \textcolor{blue}{\textbf{0.333}} & \textcolor{blue}{\textbf{0.370}} & \textcolor{purple}{\underline{0.345}} & \textcolor{purple}{\underline{0.380}} & 0.367 & 0.417 & 0.364 & 0.389 & 0.477 & 0.463 & 0.365 & 0.401 & {0.358} & {0.386} & 0.371 & 0.392 & 0.411 & 0.409 & 0.377 & 0.405 & 0.360 & 0.392 & 0.505 & 0.442 \\
ETTm2   & \textcolor{blue}{\textbf{0.252}} & \textcolor{blue}{\textbf{0.311}} & \textcolor{purple}{\underline{0.259}} & \textcolor{purple}{\underline{0.319}} & 0.281 & 0.341 & 0.277 & 0.327 & 0.310 & 0.359 & 0.283 & 0.337 & 0.268 & 0.325 & 0.281 & 0.333 & 0.277 & 0.330 & 0.282 & 0.338 & 0.284 & 0.341 & 0.293 & 0.347 \\
Weather & \textcolor{blue}{\textbf{0.223}} & \textcolor{blue}{\textbf{0.261}} & \textcolor{purple}{\underline{0.239}} & \textcolor{purple}{\underline{0.278}} & 0.255 & 0.298 & 0.252 & 0.290 & 0.271 & 0.308 & 0.248 & 0.284 & 0.247 & 0.282 & 0.322 & 0.355 & {0.244} & {0.282} & 0.258 & 0.286 & 0.247 & 0.284 & 0.260 & 0.291 \\
Electricity    & \textcolor{blue}{\textbf{0.162}} & \textcolor{blue}{\textbf{0.254}} & \textcolor{blue}{\textbf{0.162}} & \textcolor{purple}{\underline{0.255}} & 0.239 & 0.296 & 0.168 & 0.261 & 0.185 & 0.288 & 0.257 & 0.354 & \textcolor{purple}{\underline{0.167}} & 0.261 & 0.173 & 0.272 & \textcolor{purple}{\underline{0.167}} & {0.257} & \textcolor{purple}{\underline{0.167}} & 0.260 & 0.180 & 0.283 & 0.207 & 0.304 \\
Traffic & {0.383} & {0.262} & \textcolor{blue}{\textbf{0.373}} & \textcolor{blue}{\textbf{0.253}} & 0.891 & 0.442 & 0.379 & 0.265 & 0.414 & 0.305 & 0.428 & 0.312 & 0.436 & 0.317 & \textcolor{purple}{\underline{0.378}} & \textcolor{purple}{\underline{0.256}} & 0.442 & 0.321 & 0.384 & 0.272 & 0.408 & 0.298 & 0.619 & 0.330 \\
\midrule
\rowcolor{gray!15} 
\textbf{$1^{st}$ Count} & \multicolumn{2}{c|}{\textcolor{blue}{\textbf{11}}} & \multicolumn{2}{c|}{\textcolor{purple}{\underline{4}}} & \multicolumn{2}{c|}{0} & \multicolumn{2}{c|}{0} & \multicolumn{2}{c|}{0} & \multicolumn{2}{c|}{0} & \multicolumn{2}{c|}{0} & \multicolumn{2}{c|}{0} & \multicolumn{2}{c|}{0} & \multicolumn{2}{c|}{0} & \multicolumn{2}{c|}{0} & \multicolumn{2}{c}{0} \\
\bottomrule
\end{tabular}
}
\vspace{-0.3cm}
\end{table*}

\subsubsection{\textbf{LLM Sequence Modeling}}
The instruction-aware queries $\mathbf{Q}_{out}$ form a sequence of $K=N$ high-level temporal tokens that inherit the ordering of the input patches. They are fed into the frozen LLM, whose causal self-attention realizes autoregressive, next-token reasoning over the trajectory: conditioned on the tokens up to position $k$, the $k$-th hidden state captures the local dynamics and task intent needed to predict the segment that follows the $k$-th patch, $$\mathbf{H}_{llm} = \text{LLM}(\mathbf{Q}_{out}), \quad \mathbf{H}_{llm}\in\mathbb{R}^{K\times D_{llm}},$$ where $\mathbf{H}_{llm}$ collects the last-layer hidden states.

\subsubsection{\textbf{De-tokenization and Forecasting Head}}
A de-tokenization head $\Phi_{dec}$, implemented as a linear layer or an MLP, maps each hidden state back to the numerical space, producing a segment of length $S$: $$\hat{\mathbf{P}} = \Phi_{dec}(\mathbf{H}_{llm}), \quad \hat{\mathbf{P}}\in\mathbb{R}^{K\times S}.$$ By the next-token convention, the $k$-th predicted segment $\hat{\mathbf{P}}_k$ corresponds to the patch following the $k$-th input patch. During training, the $K$ segments are reshaped and concatenated into the raw forecast $\hat{\mathbf{Y}}_{raw}$ and supervised against the aligned ground-truth patches; applying inverse instance normalization~\cite{KimKTPCC22} then yields the final forecast $\hat{\mathbf{Y}}$.

\subsection{Joint Optimization Objective and Inference}
\label{sec:opt_inference}
\subsubsection{\textbf{Joint Optimization Objective}}
During the training phase, we optimize a multi-task objective that balances forecasting precision with cross-modal semantic consistency. To preserve pre-trained sequence modeling capabilities, the LLM backbone remains frozen.
The primary objective is the Mean Squared Error (MSE), minimizing the discrepancy between predicted values $\hat{\mathbf{Y}}$ and ground truth $\mathbf{Y}$:
$$\mathcal{L}_{forecast} = \frac{1}{T \times C} \sum_{t=1}^{T} \sum_{c=1}^{C} (y_{t,c} - \hat{y}_{t,c})^2$$
To bridge the modality gap, we further introduce a patch-wise contrastive objective aligning each temporal patch with its paired linguistic descriptor. Let $\mathbf{h}_i$ be the encoded representation of the $i$-th patch and $\mathbf{e}^{P}_i$ the embedding of its patch-level prompt. We use the temporal projection head $g_t$ from the TL-Connector and introduce a prompt-side head $g_p$ to map both into the shared alignment space: $$\hat{\mathbf{h}}_i = g_t(\mathbf{h}_i), \qquad \hat{\mathbf{e}}_i = g_p(\mathbf{e}^{P}_i),$$ where $\hat{\mathbf{h}}_i$ is the $i$-th row of the projected features $\hat{\mathbf{H}}$ that the queries probe, and $i$ indexes the patch--prompt pairs in a mini-batch. We then apply an InfoNCE loss~\cite{oord2018representation}:
$$\mathcal{L}_{align} = -\frac{1}{BN} \sum_{i=1}^{BN}
\log \frac{\exp(\text{sim}(\hat{\mathbf{h}}_i, \hat{\mathbf{e}}_{i}) / \tau)}
{\sum_{j=1}^{BN} \exp(\text{sim}(\hat{\mathbf{h}}_i, \hat{\mathbf{e}}_{j}) / \tau)},$$
where $\text{sim}(\cdot,\cdot)$ is cosine similarity and $\tau$ a temperature. By pulling
matched pairs together and pushing apart mismatched ones, it forces the encoder to embed
each temporal pattern close to its semantic descriptor.
Note that patch-level prompts are utilized exclusively during training to supervise the encoder. The joint optimization objective is defined as:
$$\mathcal{L}_{total} = \mathcal{L}_{forecast} + \alpha \mathcal{L}_{align},$$
where $\alpha$ is a weighting coefficient.

\subsubsection{\textbf{Model Inference}}
During inference, patch-level prompts are discarded and the model relies solely on the global task instruction and the learned probing mechanism. Forecasting proceeds autoregressively: a single forward pass yields the next segment of length $S$, which is appended to the lookback window while the oldest observations are dropped. Repeating this for $\lceil T/S\rceil$ steps produces the full horizon $T$, allowing \abb to stay sensitive to evolving signal trends throughout the forecast.

\begin{table*}[h]
\caption{Multivariate long-term forecasting results under one-for-one protocol. Our single model trained only on a 96-step horizon is compared against baseline models that have been independently trained and optimized for each specific prediction length. \textcolor{blue}{\textbf{Blue}} and \textcolor{purple}{\underline{Purple}} denote the best and second-best results, respectively. Full results are provided in Appendix.}
\label{tab:Forecast Results One-for-one}
\centering
\renewcommand{\arraystretch}{1.3}
\setlength{\tabcolsep}{8.5pt}
\resizebox{\textwidth}{!}{
\begin{tabular}{l | cc | cc | cc | cc | cc | cc | cc | cc | cc}
\toprule
\multirow{3}{*}{\textbf{Model}} & \multicolumn{2}{c|}{\textbf{One-for-all}} & \multicolumn{16}{c}{\textbf{Trained respectively on specific prediction lengths (One-for-one)}} \\
\cmidrule(lr){2-3} \cmidrule(lr){4-19}
& \multicolumn{2}{c|}{{\abb}} & \multicolumn{2}{c|}{CALF} & \multicolumn{2}{c|}{TimeLLM} & \multicolumn{2}{c|}{FPT} & \multicolumn{2}{c|}{SimpleTM} & \multicolumn{2}{c|}{TimeMixer} & \multicolumn{2}{c|}{iTransformer} & \multicolumn{2}{c|}{PatchTST} & \multicolumn{2}{c}{TimesNet} \\
& \multicolumn{2}{c|}{{(Ours)}} & \multicolumn{2}{c|}{\cite{liu2025calf}} & \multicolumn{2}{c|}{\cite{0005WMCZSCLLPW24}} & \multicolumn{2}{c|}{\cite{zhou2023one}} & \multicolumn{2}{c|}{\cite{ChenLMS25}} & \multicolumn{2}{c|}{\cite{WangWSHLMZ024}} & \multicolumn{2}{c|}{\cite{LiuHZWWML24}} & \multicolumn{2}{c|}{\cite{NieNSK23}} & \multicolumn{2}{c}{\cite{WuHLZ0L23}} \\
\midrule
Metric & MSE & MAE & MSE & MAE & MSE & MAE & MSE & MAE & MSE & MAE & MSE & MAE & MSE & MAE & MSE & MAE & MSE & MAE \\
\midrule
ETTh1   & \textcolor{blue}{\textbf{0.395}} & \textcolor{blue}{\textbf{0.417}} & 0.440 & 0.452 & 0.578 & 0.529 & 0.438 & 0.446 & \textcolor{purple}{\underline{0.422}} & 0.449 & 0.428 & \textcolor{purple}{\underline{0.442}} & 0.451 & 0.465 & 0.468 & 0.467 & 0.484 & 0.489 \\
ETTh2   & \textcolor{blue}{\textbf{0.338}} & \textcolor{blue}{\textbf{0.386}} & 0.366 & 0.402 & 0.435 & 0.455 & 0.396 & 0.427 & \textcolor{purple}{\underline{0.361}} & \textcolor{purple}{\underline{0.395}} & 0.374 & 0.409 & 0.400 & 0.426 & 0.417 & 0.438 & 0.433 & 0.455 \\
ETTm1   & \textcolor{blue}{\textbf{0.333}} & \textcolor{blue}{\textbf{0.370}} & 0.363 & 0.393 & 0.406 & 0.417 & 0.359 & \textcolor{purple}{\underline{0.390}} & \textcolor{purple}{\underline{0.356}} & \textcolor{purple}{\underline{0.390}} & 0.418 & 0.423 & 0.372 & 0.403 & 0.387 & 0.409 & 0.444 & 0.434 \\
ETTm2   & \textcolor{blue}{\textbf{0.252}} & \textcolor{blue}{\textbf{0.311}} & \textcolor{purple}{\underline{0.266}} & \textcolor{purple}{\underline{0.321}} & 0.290 & 0.345 & 0.274 & 0.330 & 0.269 & 0.329 & 0.269 & 0.327 & 0.274 & 0.335 & 0.289 & 0.343 & 0.303 & 0.353 \\
Weather & \textcolor{blue}{\textbf{0.223}} & \textcolor{blue}{\textbf{0.261}} & 0.241 & 0.281 & 0.273 & 0.313 & 0.242 & 0.282 & 0.244 & 0.281 & 0.262 & 0.293 & 0.261 & 0.290 & \textcolor{purple}{\underline{0.240}} & \textcolor{purple}{\underline{0.280}} & 0.252 & 0.290 \\
Electricity     & \textcolor{blue}{\textbf{0.162}} & \textcolor{blue}{\textbf{0.254}} & 0.165 & 0.262 & 0.176 & 0.276 & 0.166 & 0.263 & 0.166 & 0.261 & 0.166 & \textcolor{purple}{\underline{0.257}} & \textcolor{purple}{\underline{0.163}} & 0.258 & 0.166 & 0.267 & 0.203 & 0.307 \\
Traffic & \textcolor{blue}{\textbf{0.383}} & \textcolor{blue}{\textbf{0.262}} & \textcolor{purple}{\underline{0.386}} & \textcolor{purple}{\underline{0.265}} & 0.402 & 0.284 & 0.408 & 0.288 & 0.453 & 0.331 & 0.404 & 0.285 & 0.386 & 0.275 & 0.397 & 0.279 & 0.622 & 0.329 \\
\midrule
\rowcolor{gray!15} 
\textbf{$1^{st}$ Count} & \multicolumn{2}{c|}{\textcolor{blue}{\textbf{14}}} & \multicolumn{2}{c|}{0} & \multicolumn{2}{c|}{0} & \multicolumn{2}{c|}{0} & \multicolumn{2}{c|}{0} & \multicolumn{2}{c|}{0} & \multicolumn{2}{c|}{0} & \multicolumn{2}{c|}{0} & \multicolumn{2}{c}{0} \\
\bottomrule
\end{tabular}
}
\vspace{-0.3cm}
\end{table*}

\section{Experiments}
\subsection{Experimental Setup}
\subsubsection{\textbf{Datasets}}
We evaluate our model on seven widely recognized real-world benchmarks for long-term time series forecasting: ETT (ETTh1, ETTh2, ETTm1, ETTm2), Electricity, Weather, and Traffic \cite{WuHLZ0L23}. These datasets span diverse domains, including power grid management, meteorology, and transportation, offering a comprehensive test for the model's robustness and generalization. Detailed statistics are given in the Appendix. 

\subsubsection{\textbf{Competitive Baselines}}
To demonstrate the superiority, we compare \abb against two categories of methods. LLM-based Forecasters: TALON \cite{sun2025adapting}, AutoTimes \cite{liu2024autotimes}, Time-LLM \cite{0005WMCZSCLLPW24}, and FPT \cite{zhou2023one}. Deep Learning forecasters: SimpleTM \cite{ChenLMS25}, Timer-XL \cite{LiuQ00L25}, TimeMixer \cite{WangWSHLMZ024}, iTransformer \cite{LiuHZWWML24}, PatchTST \cite{NieNSK23}, and TimesNet \cite{WuHLZ0L23}.

\subsubsection{\textbf{Implementation Details}}
We follow the experimental protocols of recent LLM-based forecasters~\cite{sun2025adapting,liu2024autotimes} for a fair comparison, and adopt the same lookback window of $L = 672$ across all methods. The LLM backbone is a pre-trained, frozen GPT-2~\cite{radford2019language}, and the temporal encoder follows the PatchTST architecture~\cite{NieNSK23} to provide the initial patch embeddings. For baselines, we report the results under the identical protocol from~\cite{sun2025adapting}. \abb is trained and evaluated under the same setting. All experiments use NVIDIA A100 (80GB) GPUs to ensure sufficient computational throughput.

\subsection{Time Series Forecasting Comparative Analysis}
\subsubsection{\textbf{Evaluation Protocols}} 
We assess predictive performance under two distinct protocols. \textit{(1) One-for-all}: a single model is trained only on the 96-step horizon and applied to all longer horizons by recursively feeding its predictions back into the lookback window (the rolling-forecast setting). \textit{(2) One-for-one}: following the standard protocol~\cite{WuHLZ0L23}, a separate model is trained for each prediction length $\{96, 192, 336, 720\}$. The first protocol is the more demanding of the two, since recursive generation accumulates error over the horizon.

\subsubsection{\textbf{Results under One-for-all Protocol}} 
As shown in Table~\ref{tab:Rolling Forecast Results}, \abb outperforms both LLM-based and deep-learning baselines in the rolling-forecast setting, achieving the best score on 11 of the 14 averaged metrics across the seven benchmarks (full per-horizon results in the Appendix). The gains are most pronounced on the minute-level ETTm2 and the multivariate Weather and Electricity datasets, where non-stationary, high-frequency dynamics are prevalent. Because errors compound under recursion, this margin suggests that instruction conditioning helps the queries track the evolving signal rather than drift.

\subsubsection{\textbf{Results under One-for-one Protocol}} 
Table~\ref{tab:Forecast Results One-for-one} compares our single 96-step model against baselines trained separately for each horizon. \abb attains the best result on every averaged metric, despite using a fraction of the training cost. A single model serves all horizons, whereas iTransformer and PatchTST require a dedicated run per length. That one model matches or surpasses horizon-specific training indicates it captures horizon-agnostic temporal structure under instruction guidance, rather than fitting a particular forecast length.

\subsection{Zero-shot Forecasting Comparative Analysis}
\subsubsection{\textbf{Evaluation Protocols}} 
To evaluate the transferability and robustness of \abb, we conduct zero-shot forecasting experiments following the benchmark protocol. The model is trained on a specific source domain and subsequently evaluated on an entirely unseen target domain without any further parameter updates. We utilize the ETT datasets to examine cross-domain performance across varying temporal resolutions and differing system dynamics.

\subsubsection{\textbf{Results and Analysis}} 
Table \ref{tab:Zero-shot Results_Updated} summarizes the zero-shot performance across 12 transfer scenarios. The empirical evidence underscores the superior generalizability of our model. We achieve the best performance in 25 out of 32 individual metric comparisons, significantly outperforming established baselines. Notably, in challenging scenarios such as $h1 \rightarrow m1$ and $h2 \rightarrow m1$, our model reduces the MSE by approximately 30\% to 37\% compared to the strongest baseline. Performance also remains stable across changes in sampling frequency, indicating that the instruction-aware probes transfer the relevant temporal semantics rather than overfitting the source statistics.

\subsection{Ablation Studies.}
\label{sec:ablation}
In this section, we conduct a series of ablation experiments to quantify the contribution of each core component in \abb. By systematically removing or replacing key modules, we demonstrate the necessity of our architectural design for effective instruction-aware forecasting.

\begin{table}[t]
\caption{Zero-shot forecasting performance across diverse ETT domain transfers. \textcolor{blue}{\textbf{Blue}} and \textcolor{purple}{\underline{Purple}} denote the best and second-best results, respectively. Full results are provided in Appendix.}
\label{tab:Zero-shot Results_Updated}
\centering
\renewcommand{\arraystretch}{1.3}
\setlength{\tabcolsep}{8.5pt}
\resizebox{\linewidth}{!}{
\begin{tabular}{l | cc | cc | cc | cc}
\toprule
\multirow{2}{*}{\textbf{Models}} & \multicolumn{2}{c|}{{\abb}} & \multicolumn{2}{c|}{TALON} & \multicolumn{2}{c|}{AutoTimes} & \multicolumn{2}{c}{Timer\_XL} \\
& \multicolumn{2}{c|}{{(Ours)}} & \multicolumn{2}{c|}{\cite{sun2025adapting}} & \multicolumn{2}{c|}{\cite{liu2024autotimes}} & \multicolumn{2}{c}{\cite{LiuQ00L25}} \\
\midrule
Metric & MSE & MAE & MSE & MAE & MSE & MAE & MSE & MAE \\
\midrule
h1$\rightarrow$h2 & 0.361 & \textcolor{blue}{\textbf{0.395}} & \textcolor{blue}{\textbf{0.357}} & \textcolor{purple}{\underline{0.396}} & \textcolor{purple}{\underline{0.360}} & 0.399 & 0.373 & 0.408 \\
h1$\rightarrow$m1 & \textcolor{blue}{\textbf{0.477}} & \textcolor{blue}{\textbf{0.448}} & \textcolor{purple}{\underline{0.760}} & \textcolor{purple}{\underline{0.572}} & 0.815 & \textcolor{purple}{\underline{0.572}} & 0.820 & 0.591 \\
h1$\rightarrow$m2 & \textcolor{blue}{\textbf{0.279}} & \textcolor{blue}{\textbf{0.331}} & \textcolor{purple}{\underline{0.317}} & \textcolor{purple}{\underline{0.368}} & 0.343 & 0.383 & \textcolor{purple}{\underline{0.343}} & 0.383 \\
\rowcolor{gray!10} Avg. & \textcolor{blue}{\textbf{0.372}} & \textcolor{blue}{\textbf{0.391}} & \textcolor{purple}{\underline{0.478}} & \textcolor{purple}{\underline{0.445}} & 0.506 & 0.451 & 0.512 & 0.461 \\
\midrule
h2$\rightarrow$h1 & \textcolor{blue}{\textbf{0.497}} & \textcolor{blue}{\textbf{0.483}} & \textcolor{purple}{\underline{0.580}} & \textcolor{purple}{\underline{0.536}} & 0.741 & 0.608 & 0.583 & 0.542 \\
h2$\rightarrow$m1 & \textcolor{blue}{\textbf{0.550}} & \textcolor{blue}{\textbf{0.477}} & \textcolor{purple}{\underline{0.772}} & \textcolor{purple}{\underline{0.582}} & 1.067 & 0.655 & 0.852 & 0.625 \\
h2$\rightarrow$m2 & \textcolor{blue}{\textbf{0.281}} & \textcolor{blue}{\textbf{0.334}} & \textcolor{purple}{\underline{0.311}} & \textcolor{purple}{\underline{0.362}} & 0.328 & 0.377 & 0.340 & 0.377 \\
\rowcolor{gray!10} Avg. & \textcolor{blue}{\textbf{0.443}} & \textcolor{blue}{\textbf{0.431}} & \textcolor{purple}{\underline{0.554}} & \textcolor{purple}{\underline{0.493}} & 0.712 & 0.547 & 0.592 & 0.515 \\
\midrule
m1$\rightarrow$h1 & \textcolor{blue}{\textbf{0.487}} & \textcolor{blue}{\textbf{0.473}} & 0.624 & 0.546 & \textcolor{purple}{\underline{0.605}} & \textcolor{purple}{\underline{0.531}} & 0.733 & 0.596 \\
m1$\rightarrow$h2 & \textcolor{blue}{\textbf{0.393}} & \textcolor{blue}{\textbf{0.417}} & \textcolor{blue}{\textbf{0.393}} & \textcolor{purple}{\underline{0.426}} & 0.412 & 0.433 & 0.405 & 0.433 \\
m1$\rightarrow$m2 & \textcolor{blue}{\textbf{0.264}} & \textcolor{blue}{\textbf{0.318}} & \textcolor{purple}{\underline{0.279}} & \textcolor{purple}{\underline{0.331}} & 0.290 & 0.334 & 0.301 & 0.344 \\
\rowcolor{gray!10} Avg. & \textcolor{blue}{\textbf{0.383}} & \textcolor{blue}{\textbf{0.402}} & \textcolor{purple}{\underline{0.432}} & 0.434 & 0.436 & \textcolor{purple}{\underline{0.433}} & 0.480 & 0.458 \\
\midrule
m2$\rightarrow$h1 & \textcolor{blue}{\textbf{0.551}} & \textcolor{blue}{\textbf{0.520}} & \textcolor{purple}{\underline{0.563}} & \textcolor{blue}{\textbf{0.520}} & 0.688 & 0.568 & 0.588 & \textcolor{purple}{\underline{0.530}} \\
m2$\rightarrow$h2 & 0.371 & \textcolor{purple}{\underline{0.401}} & \textcolor{blue}{\textbf{0.356}} & \textcolor{blue}{\textbf{0.399}} & 0.380 & 0.418 & \textcolor{purple}{\underline{0.368}} & 0.408 \\
m2$\rightarrow$m1 & \textcolor{purple}{\underline{0.475}} & \textcolor{purple}{\underline{0.450}} & \textcolor{blue}{\textbf{0.452}} & \textcolor{blue}{\textbf{0.448}} & 0.490 & 0.452 & 0.525 & 0.473 \\
\rowcolor{gray!10} Avg. & \textcolor{purple}{\underline{0.466}} & \textcolor{purple}{\underline{0.457}} & \textcolor{blue}{\textbf{0.457}} & \textcolor{blue}{\textbf{0.456}} & 0.519 & 0.479 & 0.494 & 0.470 \\
\midrule
\rowcolor{gray!15} 
\textbf{1st Count} & \multicolumn{2}{c|}{\textcolor{blue}{\textbf{25}}} & \multicolumn{2}{c|}{\textcolor{purple}{\underline{9}}} & \multicolumn{2}{c|}{0} & \multicolumn{2}{c}{0} \\
\bottomrule
\end{tabular}
}
% \vspace{-0.4cm}
\end{table}

\begin{table}[t]
\centering
\caption{Abridged general ablation study results. Full ablation results are provided in Appendix.}
\label{tab:ablation_main_updated}
\renewcommand{\arraystretch}{1.2}
\setlength{\tabcolsep}{6.5pt} 
\resizebox{\columnwidth}{!}{
\begin{tabular}{l | cc | cc | cc | cc}
\toprule
\multirow{2}{*}{\textbf{Method}} & \multicolumn{2}{c|}{\textbf{ETTh1}} & \multicolumn{2}{c|}{\textbf{ETTh2}} & \multicolumn{2}{c|}{\textbf{ETTm1}} & \multicolumn{2}{c}{\textbf{ETTm2}} \\
\cmidrule(lr){2-3} \cmidrule(lr){4-5} \cmidrule(lr){6-7} \cmidrule(lr){8-9}
& MSE & MAE & MSE & MAE & MSE & MAE & MSE & MAE \\
\midrule
\textit{w/o AQG} & 0.414 & 0.433 & 0.385 & 0.416 & 0.344 & 0.374 & 0.261 & 0.317 \\
\textit{w/o TLC}   & 0.398 & 0.421 & 0.359 & 0.397 & 0.390 & 0.394 & 0.308 & 0.339 \\
\textit{w/o GTI}   & 0.397 & 0.423 & 0.359 & 0.397 & 0.342 & 0.375 & 0.259 & 0.316 \\
\textit{w/o CAL}   & 0.396 & 0.420 & 0.369 & 0.400 & 0.341 & 0.374 & 0.252 & 0.312 \\
\textit{w/o LLM}   & 0.427 & 0.438 & 0.380 & 0.401 & 0.335 & 0.370 & 0.271 & 0.321 \\
\rowcolor{gray!15}
\textbf{Ours} & \textcolor{blue}{\textbf{0.395}} & \textcolor{blue}{\textbf{0.417}} & \textcolor{blue}{\textbf{0.338}} & \textcolor{blue}{\textbf{0.386}} & \textcolor{blue}{\textbf{0.333}} & \textcolor{blue}{\textbf{0.370}} & \textcolor{blue}{\textbf{0.252}} & \textcolor{blue}{\textbf{0.311}} \\
\bottomrule
\end{tabular}
}
% \vspace{-0.2cm}
\end{table}

\subsubsection{\textbf{Component-wise Analysis}}
Table~\ref{tab:ablation_main_updated} reports five variants: \emph{w/o AQG}, replacing the adaptive queries with a fixed (static) set; \emph{w/o TLC}, replacing the TL-Connector with a direct linear projection; \emph{w/o GTI}, dropping the global task instruction; \emph{w/o CAL}, dropping the contrastive alignment objective; and \emph{w/o LLM}, replacing the frozen backbone with a linear forecasting head. Removing the adaptive queries or the connector causes the largest degradation (on ETTh2, average MSE rises from $0.338$ to $0.385$ and $0.359$, respectively), indicating that sample-specific, instruction-conditioned probing, rather than static queries or passive projection, is what drives the gains under non-stationary dynamics. Dropping the global instruction or the contrastive objective produces smaller but consistent drops, consistent with their roles as the semantic anchor and the cross-modal grounding signal. Finally, replacing the LLM with a linear head degrades ETTh1 MSE from $0.395$ to $0.427$, indicating that the framework draws on the backbone's pre-trained priors rather than using it as a trivial projector.

\begin{table}[t]
\centering
\caption{Internal ablation study of the AQG module. Full ablation results
are provided in Appendix.}
\label{tab:adpqg_ablation_main_updated}
\renewcommand{\arraystretch}{1.2}
\setlength{\tabcolsep}{6.5pt} 
\resizebox{\columnwidth}{!}{
\begin{tabular}{l | cc | cc | cc | cc}
\toprule
\multirow{2}{*}{\textbf{Method}} & \multicolumn{2}{c|}{\textbf{ETTh1}} & \multicolumn{2}{c|}{\textbf{ETTh2}} & \multicolumn{2}{c|}{\textbf{ETTm1}} & \multicolumn{2}{c}{\textbf{ETTm2}} \\
\cmidrule(lr){2-3} \cmidrule(lr){4-5} \cmidrule(lr){6-7} \cmidrule(lr){8-9}
& MSE & MAE & MSE & MAE & MSE & MAE & MSE & MAE \\
\midrule
\textit{w/o Context} & 0.403 & 0.428 & 0.372 & 0.405 & 0.336 & 0.370 & 0.285 & 0.329 \\
\textit{w/o Proj}  & 0.402 & 0.425 & 0.359 & 0.396 & 0.351 & 0.378 & 0.252 & 0.312 \\
\textit{w/o Gate}   & 0.402 & 0.424 & 0.375 & 0.407 & 0.336 & 0.370 & 0.257 & 0.316 \\
\textit{w/o P\&C}   & 0.413 & 0.430 & 0.364 & 0.396 & 0.359 & 0.380 & 0.289 & 0.331 \\
\rowcolor{gray!15}
\textbf{Ours} & \textcolor{blue}{\textbf{0.395}} & \textcolor{blue}{\textbf{0.417}} & \textcolor{blue}{\textbf{0.338}} & \textcolor{blue}{\textbf{0.386}} & \textcolor{blue}{\textbf{0.333}} & \textcolor{blue}{\textbf{0.370}} & \textcolor{blue}{\textbf{0.252}} & \textcolor{blue}{\textbf{0.311}} \\
\bottomrule
\end{tabular}
}
\vspace{-0.4cm}
\end{table}

\subsubsection{\textbf{Ablation of AQG Module}}
Table~\ref{tab:adpqg_ablation_main_updated} isolates the three mechanisms inside AQG: \emph{w/o Context} replaces attention-pooled context with average pooling; \emph{w/o Proj} removes the FiLM-based modulation; \emph{w/o Gate} removes the instance-wise selection gate; and \emph{w/o P\&C} removes both projection and context mechanisms. Replacing the attention-pooled context is the single most damaging change (average MSE rising to $0.372$ on ETTh2 and $0.285$ on ETTm2), indicating that a coarse window summary is insufficient to steer regime-specific probing. Removing the FiLM-based modulation or the selection gate also degrades accuracy, as the queries lose their per-sample adaptivity and fall back on a generic probing pattern. The worst result comes from removing both projection and context (ETTh1 MSE $0.395\!\to\!0.413$). This confirms that they are vital for effective feature interrogation, reinforcing the AQG module as a critical engine for adaptive, instruction-driven forecasting.

\begin{figure}[t]
  % \vskip 0.2in
  \begin{center}
    \centerline{\includegraphics[width=\columnwidth]{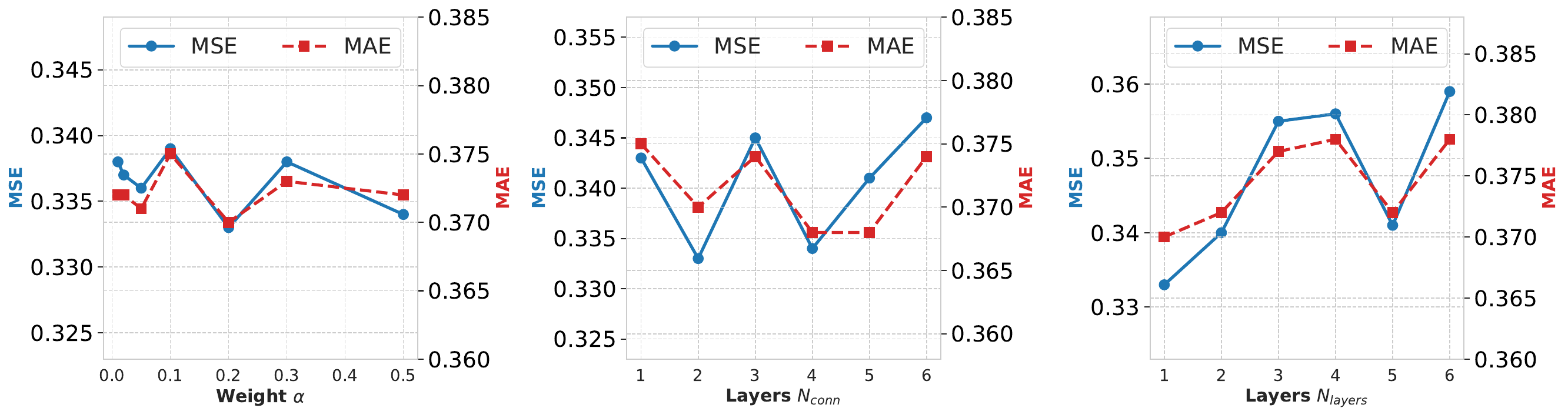}}
    \caption{
      Parameter sensitivity analysis on ETTm1 dataset. 
    }
    \label{fig:parameter_analysis}
    \vspace{-0.6cm}
  \end{center}
\end{figure}

\begin{figure}[t]
  % \vskip 0.2in
  \begin{center}
    \centerline{\includegraphics[width=\columnwidth]{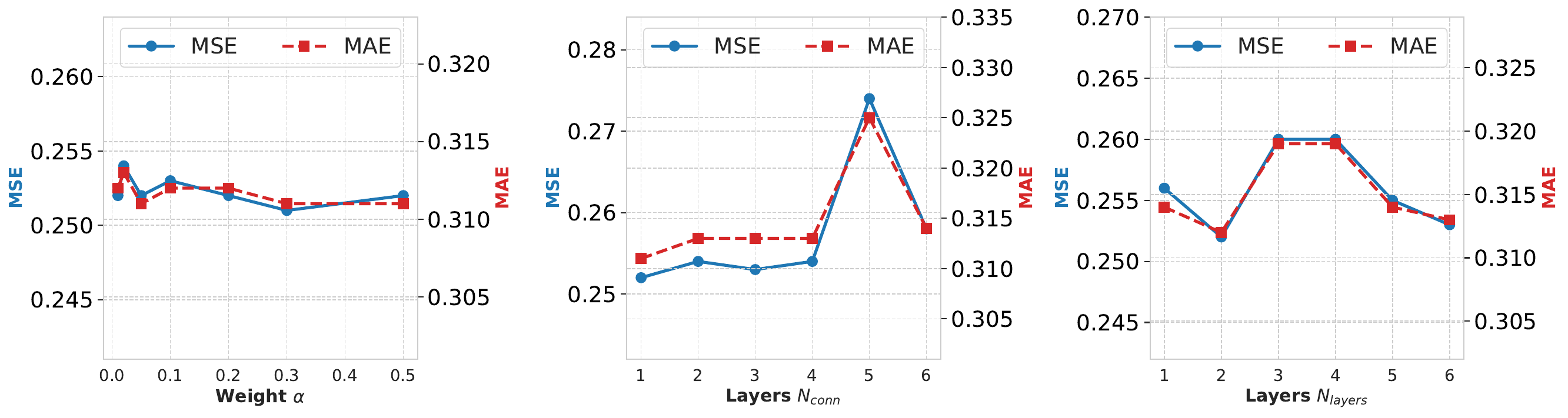}}
    \caption{
      Parameter sensitivity analysis on ETTm2 dataset. 
    }
    \label{fig:parameter_analysis_m2}
    \vspace{-0.6cm}
  \end{center}
\end{figure}

\section{Model Analysis}
In this section, we conduct parameter analysis, backbone scalability, efficiency and complexity, and visualization analysis to evaluate \abb's effectiveness. 

\subsection{Parameter Analysis}
We study \abb's sensitivity to three hyperparameters on ETTm1 and ETTm2 in Figures~\ref{fig:parameter_analysis} and \ref{fig:parameter_analysis_m2}: the contrastive weight $\alpha$, the number of connector blocks $N_{conn}$, and the number of encoder layers $N_{layers}$. Across the ranges tested, average MSE varies within a narrow band, indicating that the model is not brittle to these choices. The best results cluster around $\alpha\!=\!0.2$, $N_{conn}\!=\!2$, and $N_{layers}\!=\!1$ on ETTm1 and $\alpha\!=\!0.3$, $N_{conn}\!=\!1$, and $N_{layers}\!=\!2$ on ETTm2. Performance degrades only mildly away from this region, and a shallow connector and encoder suffice— deeper stacks bring no consistent gain. This suggests the probing mechanism extracts the salient temporal features without a deep front end, keeping the parameter and compute footprint small. Detailed results on the remaining datasets are given in the Appendix.

\subsection{{Backbone Scalability}}
\begin{figure}[t]
    \centering
    \includegraphics[width=0.9\linewidth]{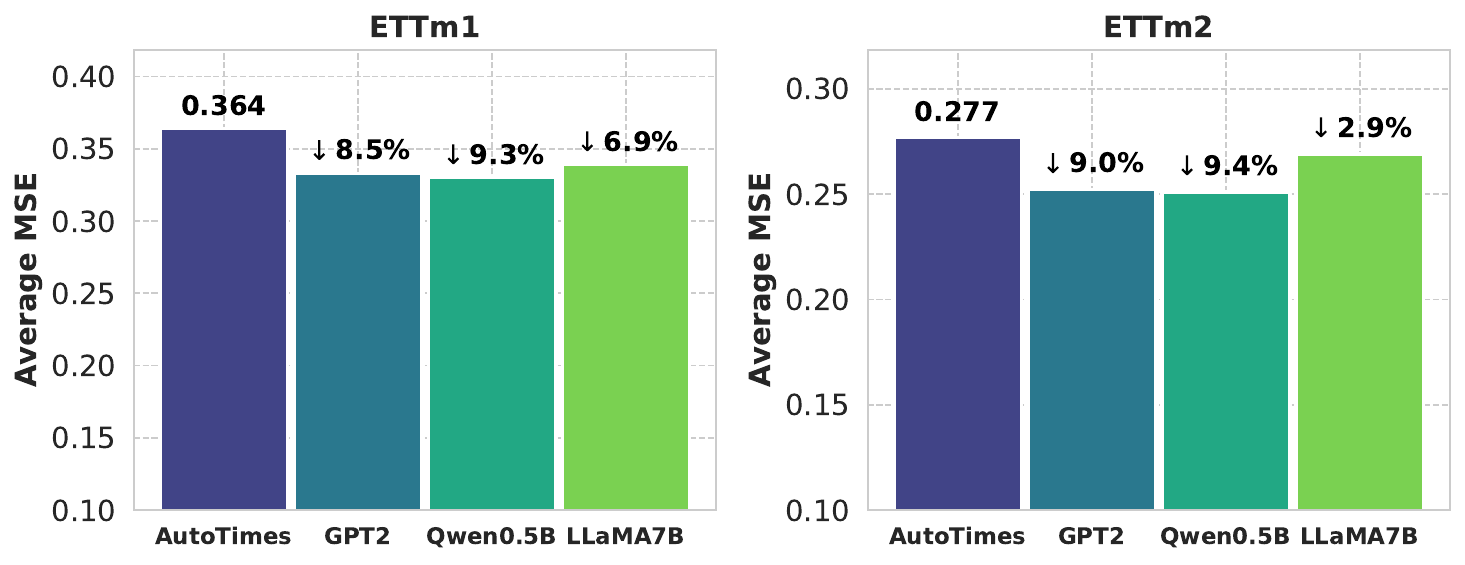}
    \caption{
      Performance across LLM backbones (GPT-2, Qwen-0.5B, LLaMA-7B) on ETTm1 and ETTm2. Percentages denote the average MSE improvement of \abb over AutoTimes under each backbone.
    }
    \label{fig:llm_backbone}
\end{figure}

To test whether the active-probing paradigm is tied to a specific backbone, we replace the default GPT-2 with Qwen-0.5B~\cite{team2024qwen2} and LLaMA-7B~\cite{touvron2023llama}, keeping all other components fixed, and compare against AutoTimes under each backbone in Figure~\ref{fig:llm_backbone}. \abb improves over AutoTimes for every backbone, confirming that the paradigm is backbone-agnostic rather than reliant on one model family. Performance does not, however, grow monotonically with scale: the Qwen-0.5B and GPT-2 variants are on par with or better than LLaMA-7B on these benchmarks, so the smallest backbones already capture the temporal structure the probes rely on. The active-probing design therefore delivers state-of-the-art accuracy at a modest backbone size, leaving the larger model as an option rather than a requirement.

\begin{figure}[t]
    \centering
    \includegraphics[width=0.65\linewidth]{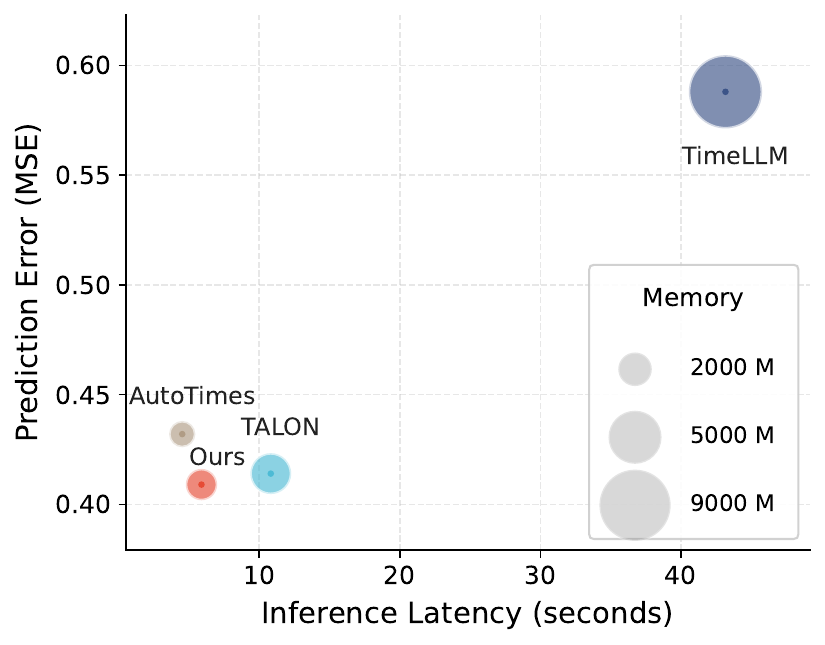}
    \caption{
      Efficiency comparison on ETTh1 (720-step horizon). The $x$-axis is inference latency, the $y$-axis is MSE, and bubble area encodes memory footprint.
    }
    \label{fig:efficacy_bubble}
\end{figure}

\begin{figure*}[t]
  % \vskip 0.2in
  \begin{center}
    \centerline{\includegraphics[width=0.95 \textwidth]{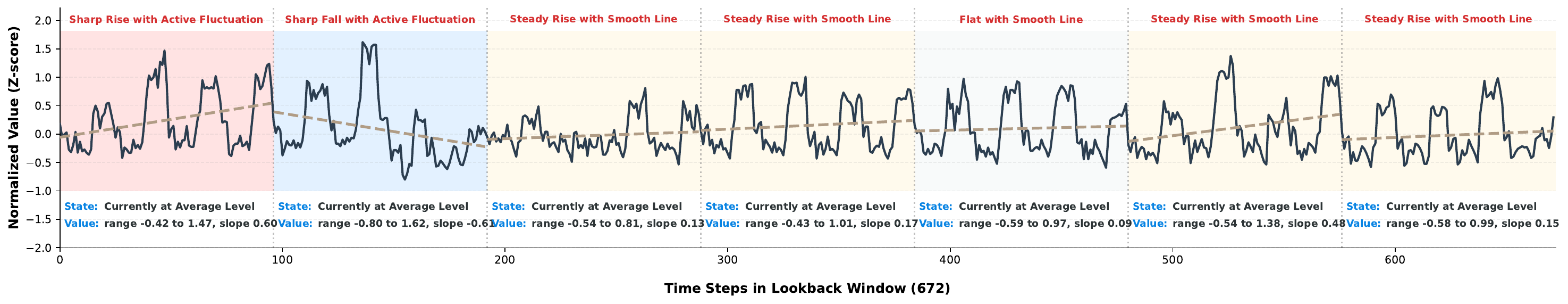}}
    \caption{
      Patch-level hybrid prompts and the corresponding time series on ETTh1 (Z-score normalized). The generated linguistic labels are consistent with the per-patch statistics (slope, range, level).
    }
    \label{fig:prompt_time_ETTh1}
  \end{center}
\end{figure*}

\begin{figure*}[t]
  % \vskip 0.2in
  \begin{center}
    \centerline{\includegraphics[width=0.95 \textwidth]{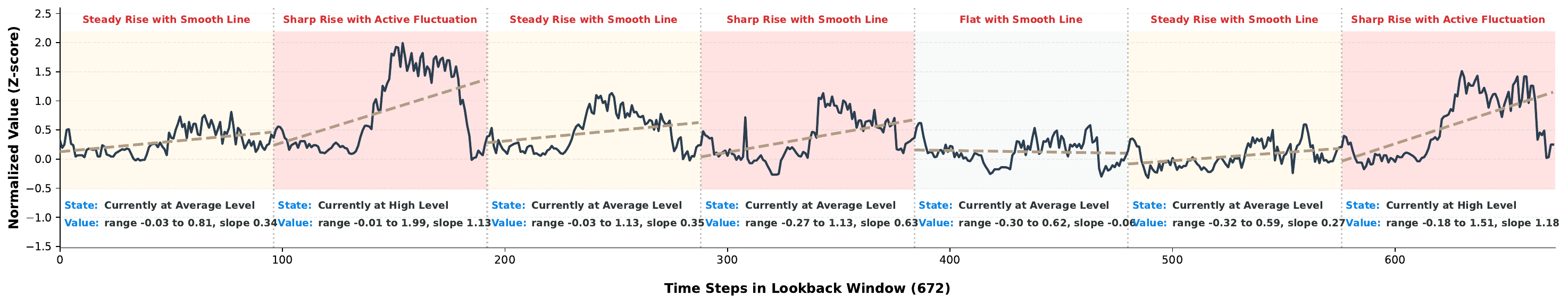}}
    \caption{
      Patch-level hybrid prompts and the corresponding time series on ETTm1 (Z-score normalized). The generated linguistic labels are consistent with the per-patch statistics (slope, range, level).
    }
    \label{fig:prompt_time_ETTm1}
  \end{center}
\end{figure*}

\subsection{{Efficiency and Complexity Analysis}}
We conduct an efficiency analysis on ETTh1 with a 720-step prediction horizon, comparing \abb against AutoTimes, TALON, and TimeLLM in terms of predictive error (MSE), inference latency, and memory footprint. As illustrated in Figure~\ref{fig:efficacy_bubble}, \abb attains the lowest prediction error among all competitors while incurring only a modest inference latency and a moderate memory footprint. In contrast, AutoTimes is marginally faster and lighter but yields noticeably higher error, whereas TALON and the LLM-based TimeLLM demand substantially greater latency and memory yet still fall short in accuracy, most strikingly, TimeLLM occupies the high-cost, high-error corner of the plot. These results demonstrate that our active probing framework delivers a more favorable trade-off between predictive precision and computational efficiency.

\subsection{Interpretability and Visualization Analysis}
\label{app:subsec:extended_interpretability_visualization}
This section provides qualitative evidence for how \abb operates, through prompt, signal alignment, attention visualizations, and forecast showcases.

\subsubsection{\textbf{Patch-level Prompts and Time Series Visualization}}
To qualitatively evaluate the effectiveness of \abb, we first examine how faithfully the generated patch-level hybrid prompts describe the underlying numerical signals on the ETTh1 and ETTm1 datasets. As illustrated in Figures \ref{fig:prompt_time_ETTh1} and \ref{fig:prompt_time_ETTm1}, our hierarchical prompts provide a precise semantic distillation of the signal dynamics within the 672-step lookback window. Specifically, the model accurately distinguishes between various temporal regimes by assigning qualitative descriptors such as \textit{``Sharp Rise with Active Fluctuation"} to volatile upward segments and \textit{``Steady Rise with Smooth Line"} to more stable trends. These semantic labels are tightly coupled with the quantitative statistics provided in the prompt, such as the local slope and value range. For instance, in the ETTh1 visualization, a segment with a steep negative slope (e.g., -0.63) is correctly identified as a \textit{``Sharp Fall"}, while sections with near-zero slopes are labeled as \textit{``Flat"}. This confirms that the instruction-construction rules map raw fluctuations to consistent semantic descriptors, which is what makes the patch prompts a usable supervision signal for the contrastive objective.

\subsubsection{\textbf{Self-Attention Visualization: Instruction Internalization}}
\begin{figure}[h] % [t] 表示置于页面顶部，figure* 必须使用 [t] 或 [p]
    \centering
    % 第一个子图
    \begin{subfigure}{\columnwidth}
        \centering
        \includegraphics[width=0.95 \columnwidth]{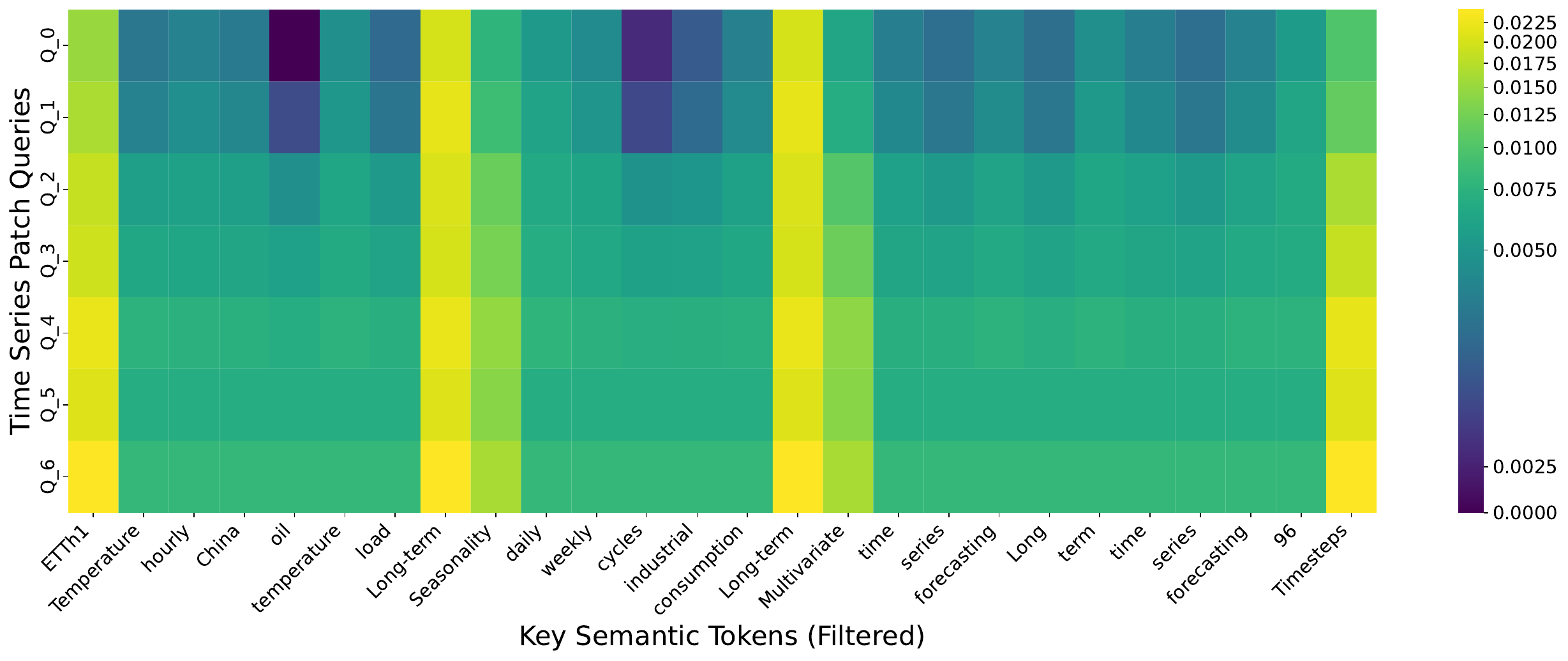}
        \caption{ETTh1}
        \label{fig:sub1}
    \end{subfigure}
    \hfill % 在子图之间填充弹性间距
    % 第二个子图
    \begin{subfigure}{\columnwidth}
        \centering
        \includegraphics[width=0.95 \columnwidth]{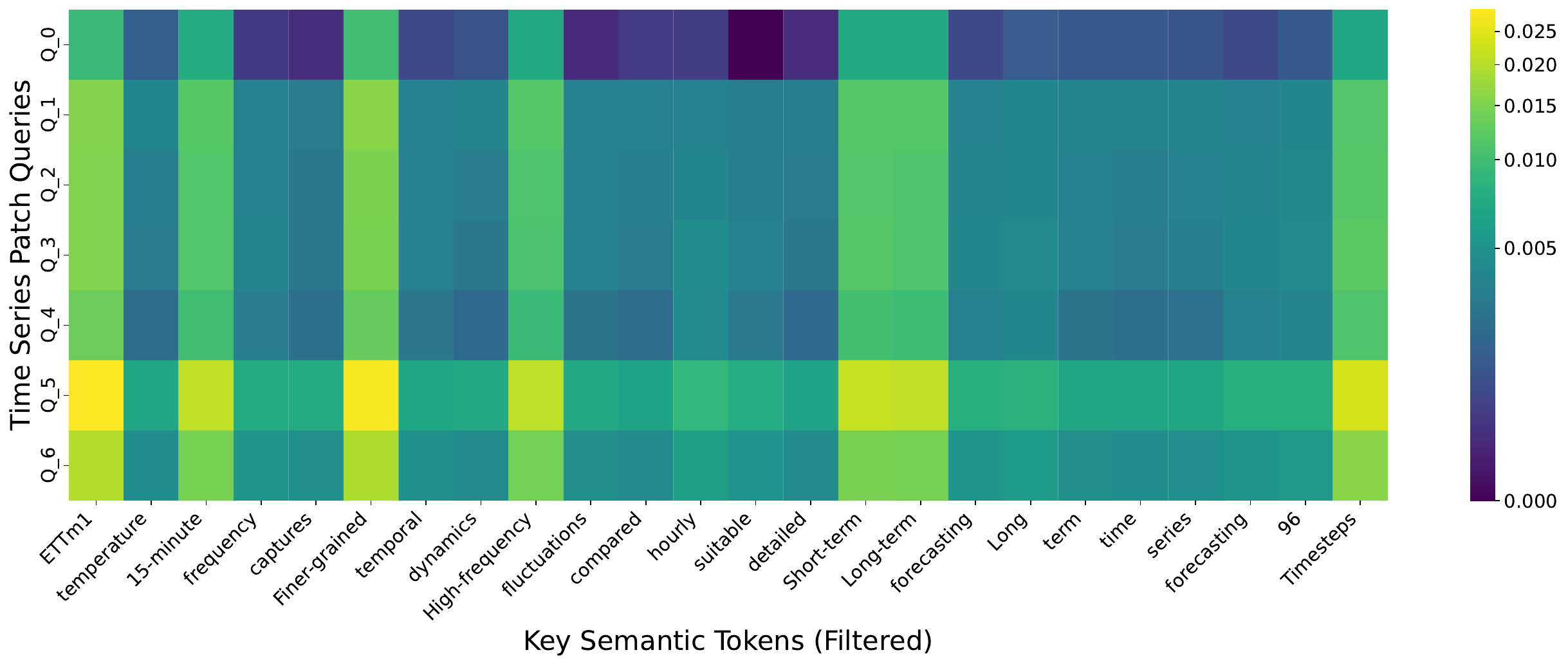}
        \caption{ETTm1}
        \label{fig:sub2}
    \end{subfigure}
    
    \caption{Self-attention visualization: Instruction internalization. Heatmaps illustrate the attention weights between adaptive queries and key filtered semantic tokens for (a) ETTh1 and (b) ETTm1, averaged across the test set. Probes place their largest weights on domain keywords (\textit{e.g., "Temperature," "15-minute"}) and task objectives, consistent with conditioning on the global task description.}
    \label{fig:self_attention}
\end{figure}

To see how the queries take up task context, we visualize the connector's self-attention weights, averaged over the test set. For readability we display the attention over content tokens, excluding padding and delimiter tokens. As depicted in the Figure \ref{fig:self_attention}, the queries exhibit distinct and concentrated attention spikes on key semantic anchors. Specifically, in the ETTh1 scenario, the probes demonstrate high sensitivity to domain-defining keywords such as \textit{``Temperature"}, \textit{``oil"}, and \textit{``Seasonality"}, alongside task constraints like \textit{``96 Timesteps"}. For the ETTm1 dataset, the attention distribution shifts significantly toward frequency-related descriptors, including \textit{``15-minute"}, \textit{``Finer-grained"}, and \textit{``High-frequency fluctuations"}. That the attended tokens change with the dataset, and do so consistently across the test set, indicates the queries are conditioned on the global task description rather than being fixed latent vectors.

\subsubsection{\textbf{Cross-Attention Visualization: Active Feature Interrogation}}
\begin{figure}[h] % [t] 表示置于页面顶部，figure* 必须使用 [t] 或 [p]
    \centering
    % 第一个子图
    \begin{subfigure}{0.48\columnwidth}
        \centering
        \includegraphics[width=1.0 \columnwidth]{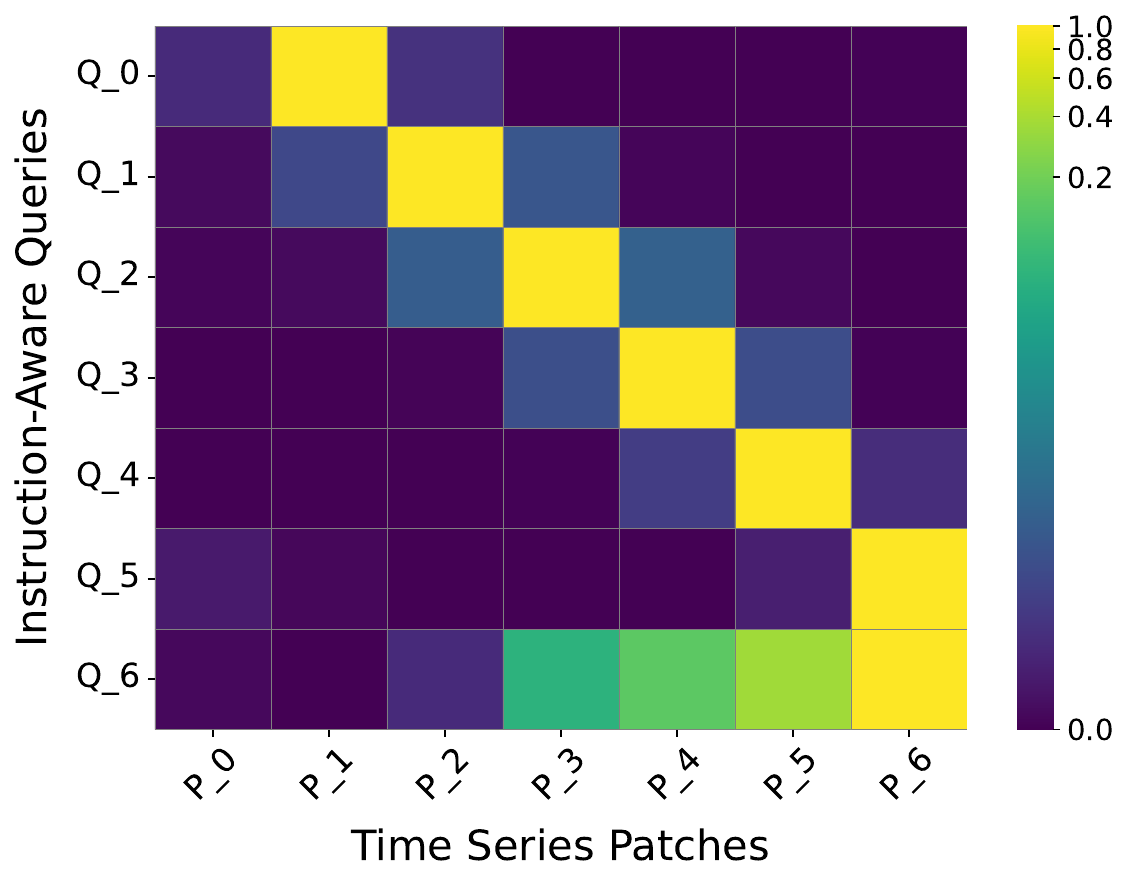}
        \caption{ETTh1}
        \label{fig:sub1}
    \end{subfigure}
    \hfill % 在子图之间填充弹性间距
    % 第二个子图
    \begin{subfigure}{0.48\columnwidth}
        \centering
        \includegraphics[width=1.0 \columnwidth]{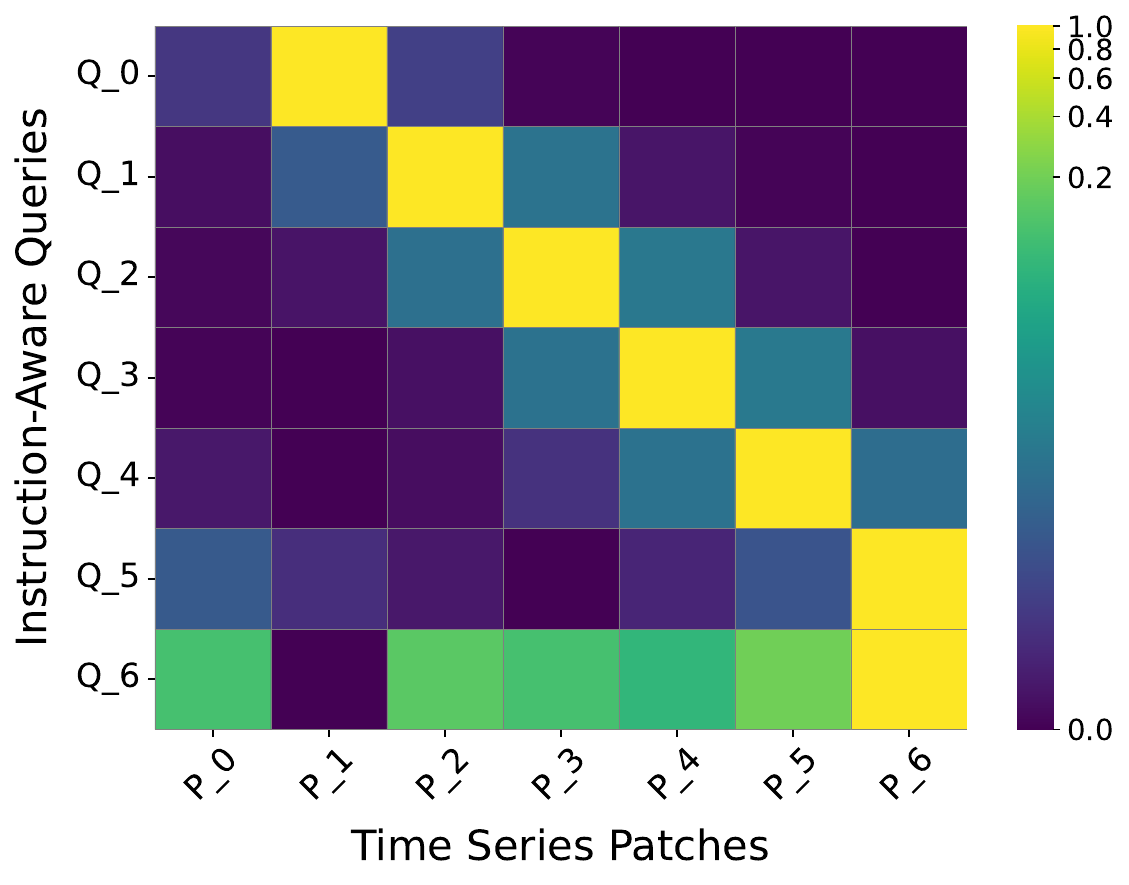}
        \caption{ETTm1}
        \label{fig:sub2}
    \end{subfigure}
    
    \caption{Cross-attention visualization: Active feature interrogation. Heatmaps show the cross-attention weights between instruction-aware queries ($Q$) and time series patches ($P$) for (a) ETTh1 and (b) ETTm1, averaged across the entire test set. The diagonal concentration indicates that specific queries learn to interrogate localized temporal segments to extract sample-specific features.}
    \label{fig:cross_attention}
\end{figure}

To further illustrate the mechanics of feature interrogation, we visualize the cross-attention weights between Instruction-aware Queries ($Q$) and temporal patches ($P$), averaged across test set on ETTh1 and ETTm1. As depicted in Figure \ref{fig:cross_attention}, the queries exhibit highly structured and localized attention patterns, effectively acting as specialized sensors that interrogate specific temporal regions within the lookback window. The observed near-diagonal pattern shows that individual queries concentrate on specific patch positions, while some queries attend more broadly. This consistent, position-selective behavior is in line with the design intent, the queries read targeted temporal regions instead of pooling all patches equally, and is what we mean by active probing.

\subsubsection{\textbf{Showcase Analysis}}
\begin{figure*}[!t] % [t] 表示置于页面顶部，figure* 必须使用 [t] 或 [p]
    \centering
    % 第一个子图
    \begin{subfigure}{0.32\textwidth}
        \centering
        \includegraphics[width=\linewidth]{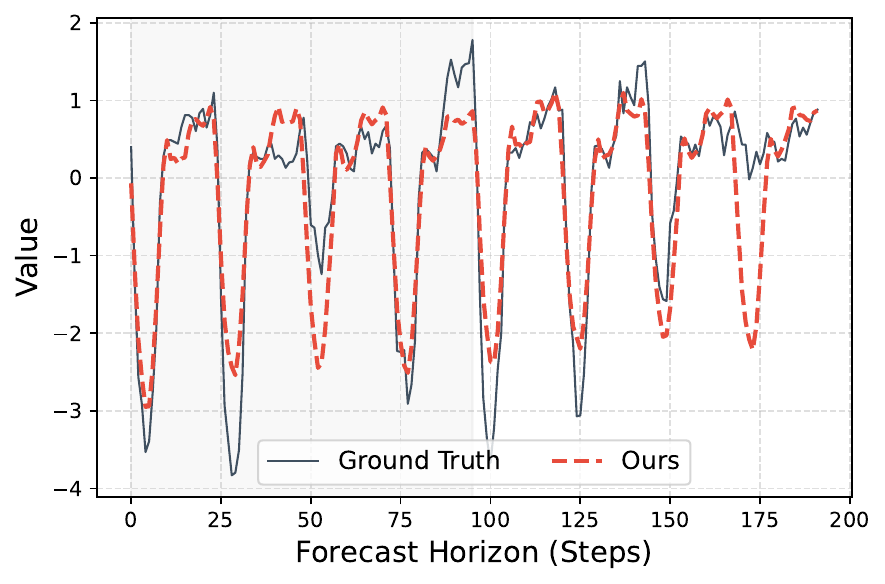}
        \caption{Ours}
        \label{fig:sub1}
    \end{subfigure}
    \hfill % 在子图之间填充弹性间距
    % 第二个子图
    \begin{subfigure}{0.32\textwidth}
        \centering
        \includegraphics[width=\linewidth]{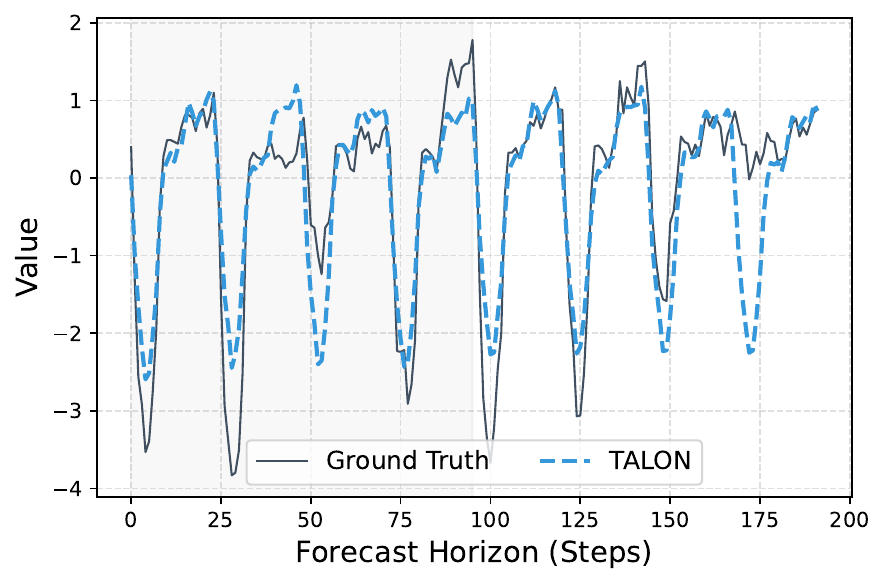}
        \caption{TALON}
        \label{fig:sub2}
    \end{subfigure}
    \hfill
    % 第三个子图
    \begin{subfigure}{0.32\textwidth}
        \centering
        \includegraphics[width=\linewidth]{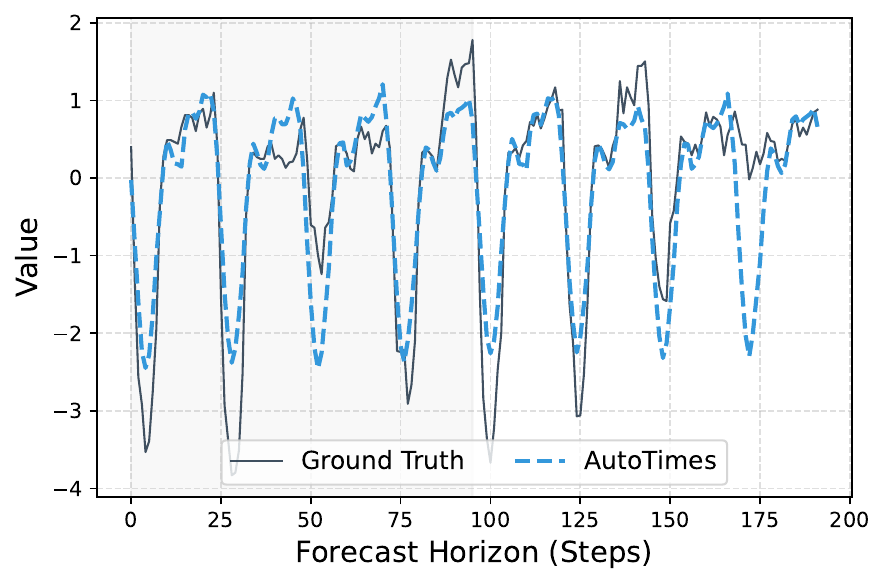}
        \caption{AutoTimes}
        \label{fig:sub3}
    \end{subfigure}
    
    \caption{Showcase forecasts on ETTh1 (192-step horizon): (a) InA-Probe, (b) TALON, (c) AutoTimes, against the ground truth.}
    \label{fig:show_case_ETTh1}
\end{figure*}

\begin{figure*}[!t] % [t] 表示置于页面顶部，figure* 必须使用 [t] 或 [p]
    \centering
    % 第一个子图
    \begin{subfigure}{0.32\textwidth}
        \centering
        \includegraphics[width=\linewidth]{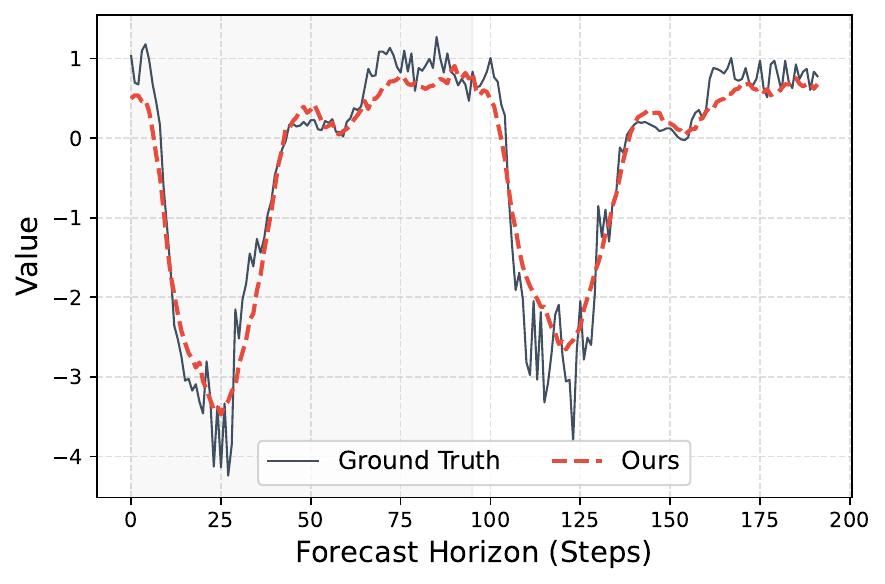}
        \caption{Ours}
        \label{fig:sub1}
    \end{subfigure}
    \hfill % 在子图之间填充弹性间距
    % 第二个子图
    \begin{subfigure}{0.32\textwidth}
        \centering
        \includegraphics[width=\linewidth]{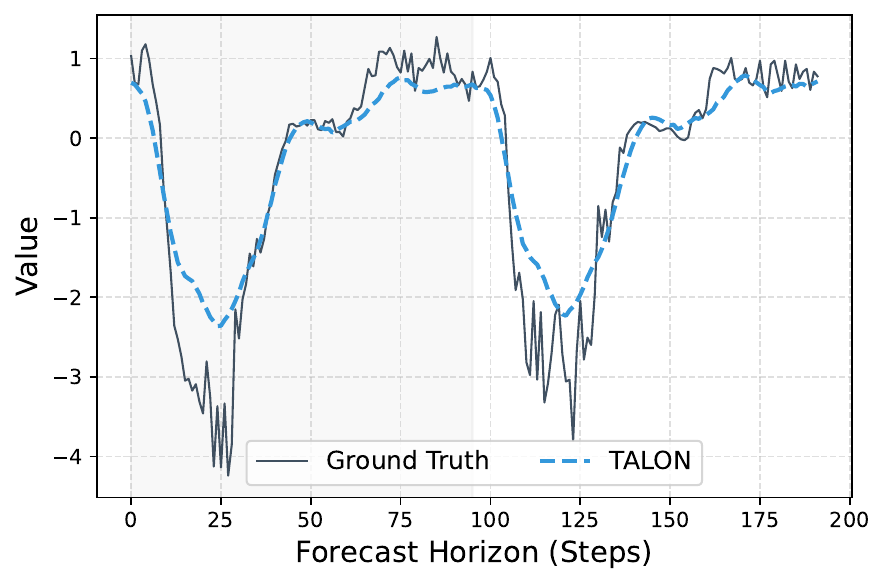}
        \caption{TALON}
        \label{fig:sub2}
    \end{subfigure}
    \hfill
    % 第三个子图
    \begin{subfigure}{0.32\textwidth}
        \centering
        \includegraphics[width=\linewidth]{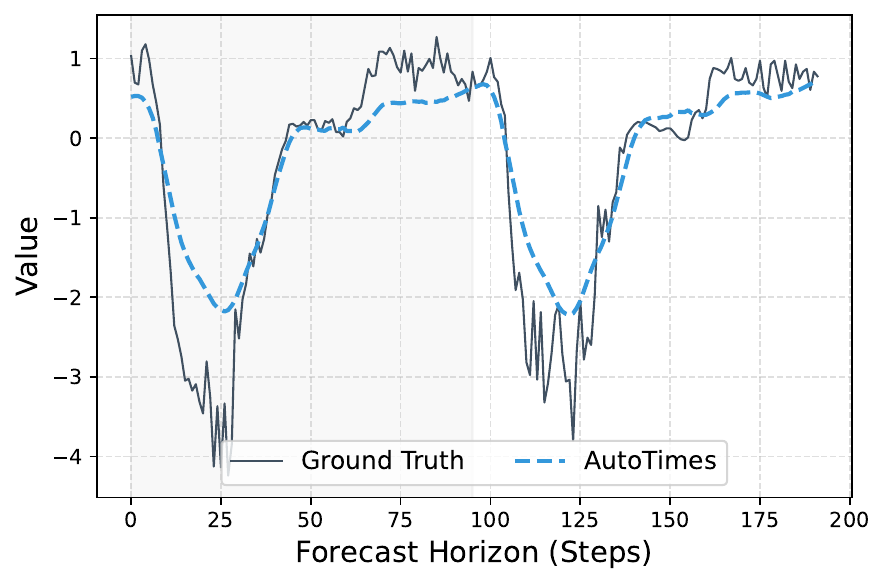}
        \caption{AutoTimes}
        \label{fig:sub3}
    \end{subfigure}
    
    \caption{Showcase forecasts on ETTm1 (192-step horizon): (a) InA-Probe, (b) TALON, (c) AutoTimes, against the ground truth.}
    \label{fig:show_case_ETTm1}
\end{figure*}

To further illustrate the forecasting quality of our framework, we show representative 192-step forecasts on ETTh1 and ETTm1 against two strong LLM-based baselines, TALON and AutoTimes.
As shown in Figure \ref{fig:show_case_ETTh1}, \abb (red dashed) tracks both the trend and the sharp peaks of the ground truth (black), whereas the baselines often struggle to capture the full amplitude of periodic fluctuations or suffer from noticeable phase lags. In the ETTm1 case shown in Figure \ref{fig:show_case_ETTm1}, \abb stays closer to the ground truth through abrupt regime changes, where the baselines tend to over-smooth. These examples are consistent with the quantitative gains and illustrate the benefit of grounding the forecast in local temporal features through the adaptive queries.

\section{Conclusion}
In this paper, we presented \abb, a framework that shifts LLM-based forecasting from passive alignment to active, instruction-aware probing. Through the AQG module and the TL-Connector, \abb generates sample-specific queries that internalize task instructions and then interrogate the encoded temporal features, with a patch-wise contrastive objective grounding those features in linguistic descriptors. Across seven benchmarks, a single \abb model matches or surpasses horizon-specific baselines under both the one-for-all and one-for-one protocols, and it transfers across temporal resolutions and domains in the zero-shot setting, reducing MSE by up to 37\% on the hardest cross-domain transfers while keeping a modest inference cost. Attention visualizations further indicate that the queries condition on the task description and read localized temporal regions rather than pooling uniformly. 
In the future, several directions remain. For example, the framework is evaluated on univariate, channel-independent forecasting, so extending it to capture cross-variable dependencies is a natural next step. Additionally, adapting the probing mechanism to online or non-stationary streaming settings would broaden its applicability. We hope \abb encourages further work on treating frozen LLMs as active reasoners over temporal signals rather than passive encoders.

\bibliography{refs}
\bibliographystyle{IEEEtran}

\end{document}